%% file: iccv19_da_seg_camera-ready.tex
\documentclass[10pt,twocolumn,letterpaper]{article}

\usepackage{iccv}
\usepackage{times}
\usepackage{epsfig}
\usepackage{graphicx}
\usepackage{amsmath}
\usepackage{amssymb}
\usepackage{soul}
\usepackage{xcolor}
\usepackage{microtype}
\usepackage{wrapfig}
\usepackage{booktabs}       
\usepackage{multirow}
\usepackage{paralist}

\usepackage[pagebackref=true,breaklinks=true,letterpaper=true,colorlinks,bookmarks=false,citecolor=blue,linkcolor=blue]{hyperref}

\iccvfinalcopy 

\def\iccvPaperID{****} 

\ificcvfinal\fi

\begin{document}
	
	\title{Domain Adaptation for Structured Output via\\Discriminative Patch Representations}
	
	\author{
		Yi-Hsuan Tsai$^{1}$
		\hspace{0.15in} Kihyuk Sohn$^1$\thanks{Now at Google Cloud AI.}
		\hspace{0.15in} Samuel Schulter$^1$ 
		\hspace{0.15in} Manmohan Chandraker$^{1,2}$ \vspace{1mm}\\
		$^1$NEC Laboratories America \hspace{0.15in} $^2$University of California, San Diego
	}
	
	\maketitle
	\ificcvfinal\thispagestyle{empty}\fi

\begin{abstract}
	Predicting structured outputs such as semantic segmentation relies on expensive per-pixel annotations to learn supervised models like convolutional neural networks.
	However, models trained on one data domain may not generalize well to other domains without annotations for model finetuning.
	To avoid the labor-intensive process of annotation, we develop a domain adaptation method to adapt the source data to the unlabeled target domain.
	We propose to learn discriminative feature representations of patches in the source domain by discovering multiple modes of patch-wise output distribution through the construction of a clustered space.
	With such representations as guidance, we use an adversarial learning scheme to push the feature representations of target patches in the clustered space closer to the distributions of source patches.
	In addition, we show that our framework is complementary to existing domain adaptation techniques and achieves consistent improvements on semantic segmentation.
	Extensive ablations and results are demonstrated on numerous benchmark datasets with various settings, such as synthetic-to-real and cross-city scenarios.
\end{abstract}

\input{intro}

\input{related}
\input{method}
\input{experiment}
\input{conclusion}

{\small
\bibliographystyle{ieee_fullname}
\bibliography{mybib}
}

\clearpage

\appendix
\input{supp}

\end{document}

%% file: intro.tex
\section{Introduction}
\label{sec:intro}

\iftrue
\begin{figure}[t]
	\centering
	\includegraphics[width=0.95\linewidth]{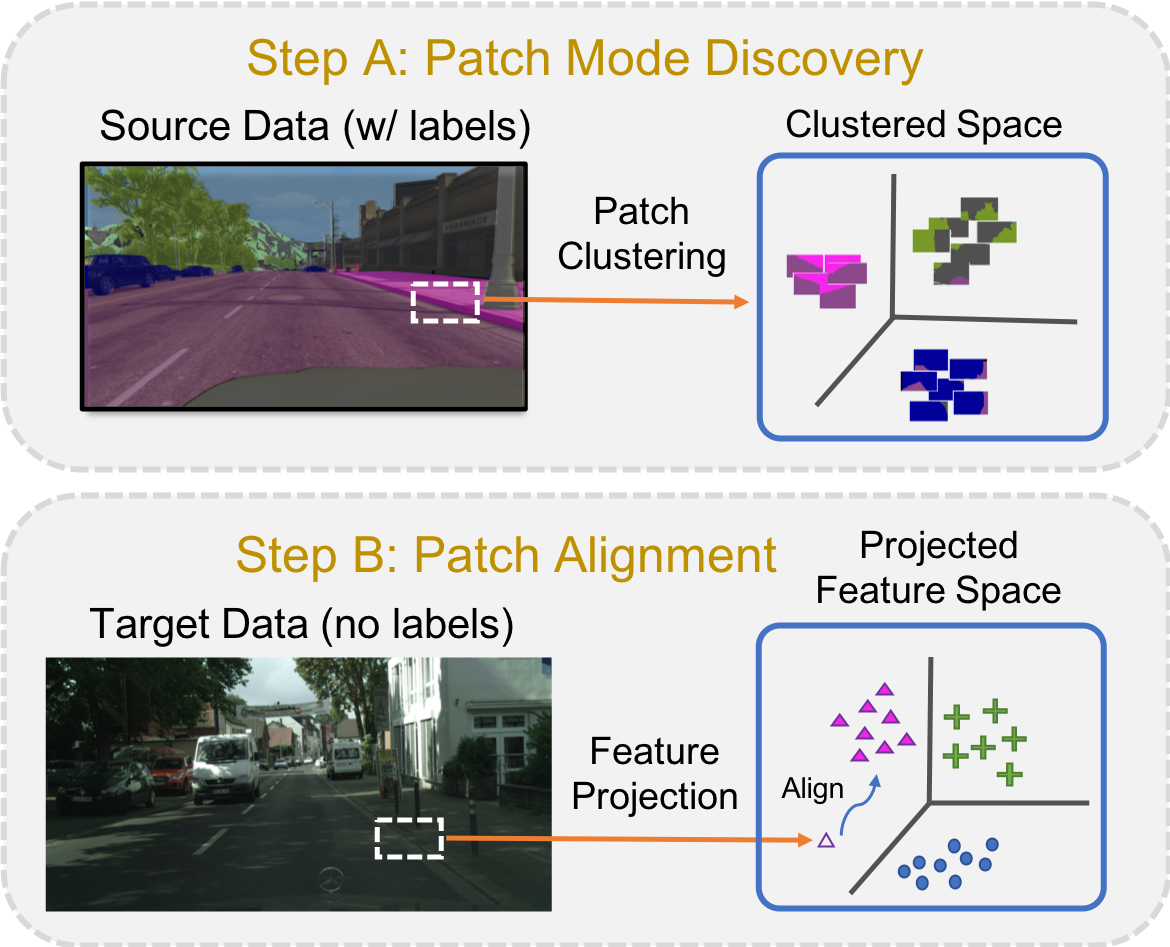}\\
	\vspace{1mm}
	\caption{
	Our method aims at improving output distribution alignment via: 1) patch mode discovery from the source patch annotations to construct a clustered space and project to a feature space, and 2) patch alignment from the target patch representation (unfilled symbol) to the source distribution (solid symbols).
	}
	\label{fig:teaser}
 	\vspace{-3mm}
\end{figure}

With the availability of large-scale annotated datasets~\cite{imagenet}, deep learning has made a significant impact on many computer vision tasks, such as object recognition~\cite{He_CVPR_2016,alexnet}, detection~\cite{girshick2015fast}, or semantic segmentation~\cite{deeplab}.
Unfortunately, learned models may not generalize when evaluated on a test domain different from the labeled training data~\cite{zhang2017understanding}.
Unsupervised domain adaptation (UDA)~\cite{ganin2016domain,saenko2010adapting} has been proposed to close the performance gap introduced by the mismatch between the source domain, where labeled data is available, and the target domain.
UDA circumvents an expensive data annotation process by utilizing only unlabeled data from the target domain.
%
Along this line, numerous UDA methods have been developed and successfully applied for classification tasks~\cite{Bousmalis_CVPR_2017,ganin2016domain,long2015learning,long2016unsupervised,saenko2010adapting,tzeng2015simultaneous,tzeng2017adversarial}.

UDA is even more crucial for pixel-level prediction tasks such as semantic segmentation as annotation is prohibitively expensive.
A prominent approach towards domain adaptation for semantic segmentation is distribution alignment by adversarial learning~\cite{Goodfellow_NIPS_2014,ganin2016domain}, where the alignment may happen at different representation layers, such as pixel-level~\cite{Hoffman_CoRR_2016,zhu2017unpaired}, feature-level~\cite{Hoffman_CoRR_2016, Huang_ECCV_2018} or output-level~\cite{Tsai_CVPR_2018}.
Despite these efforts, discovering all modes of the data distribution is a key challenge for domain adaptation~\cite{Luan_CVPR_2019}, akin to difficulties also faced by generative tasks~\cite{che2017mode,Metz_ICLR_2017}.

%
%
A critical step during adversarial training is the use of a convolutional discriminator~\cite{pix2pix,Hoffman_CoRR_2016,Tsai_CVPR_2018} that classifies patches into source or target domains.
However, the discriminator is not supervised to capture several modes in the data distribution and it may end up learning only low-level differences such as tone or texture across domains.
In addition, for the task of semantic segmentation, it is important to capture and adapt high-level patterns given the highly structured output space.
%
%
%

In this work, we propose an unsupervised domain adaptation method that explicitly discovers many modes in the structured output space of semantic segmentation to learn a better discriminator between the two domains, ultimately leading to a better domain alignment.
We leverage the pixel-level semantic annotations available in the source domain, but instead of directly working on the output space~\cite{Tsai_CVPR_2018}, our adaptation happens in two stages.
First, we extract patches from the source domain, represent them using their annotation maps and discover major modes by applying $K$-means clustering, which groups patches into $K$ clusters (Step A in Figure~\ref{fig:teaser}). Each patch in the source domain can now be assigned to a ground truth cluster/mode index.
We then introduce a $K$-way classifier that predicts the cluster/mode index of each patch, which can be supervised in the source domain but not in the target domain.

Second, different from the output space alignment~\cite{Tsai_CVPR_2018}, our method, referred as patch-level alignment (Step B in Figure~\ref{fig:teaser}) operates on the $K$-dimensional probability vector space after projecting to the clustered space that already discovers various patch modes. 
%
%
This is in contrast to prior art that operates on either pixel-~\cite{zhu2017unpaired}, feature-~\cite{Hoffman_CoRR_2016} or output-level~\cite{Tsai_CVPR_2018}.
The learned discriminator on the clustered space can back-propagate the gradient through the cluster/mode index classifier to the semantic segmentation network.
%

In experiments, we follow the setting of~\cite{Hoffman_CoRR_2016} and perform pixel-level road-scene semantic segmentation.
We experiment under various settings, including synthetic-to-real (GTA5~\cite{Richter_ECCV_2016}, SYNTHIA~\cite{Ros_CVPR_2016} to Cityscapes~\cite{cityscapes}) and cross-city (Cityscapes to Oxford RobotCar~\cite{oxford_robot}) adaptation.
We provide an extensive ablation study to validate each component in the proposed framework.
Our approach is also complementary to existing domain adaptation techniques, which we demonstrate by combining with output space adaptation~\cite{Tsai_CVPR_2018}, pixel-level adaptation~\cite{Hoffman_ICML_2018} and pseudo label re-training~\cite{Zou_ECCV_2018}.
Our results show that the learned representations improve segmentation results consistently and achieve state-of-the-art performance.
%
%
%

Our contributions are summarized as follows.
First, we propose an adversarial adaptation framework for structured prediction that explicitly tries to discover and predict modes of the output patches.
Second, we demonstrate the complementary nature of our approach by integration into three existing domain adaptation methods, which can all benefit from it.
Third, we extensively analyze our approach and show state-of-the-art results on various domain adaptation benchmarks for semantic segmentation.\footnote{The project page is at \url{https://www.nec-labs.com/~mas/adapt-seg/adapt-seg.html}.}
\fi

%% file: related.tex
\begin{figure*}[t]
	\centering
	\includegraphics[width=0.8\linewidth]{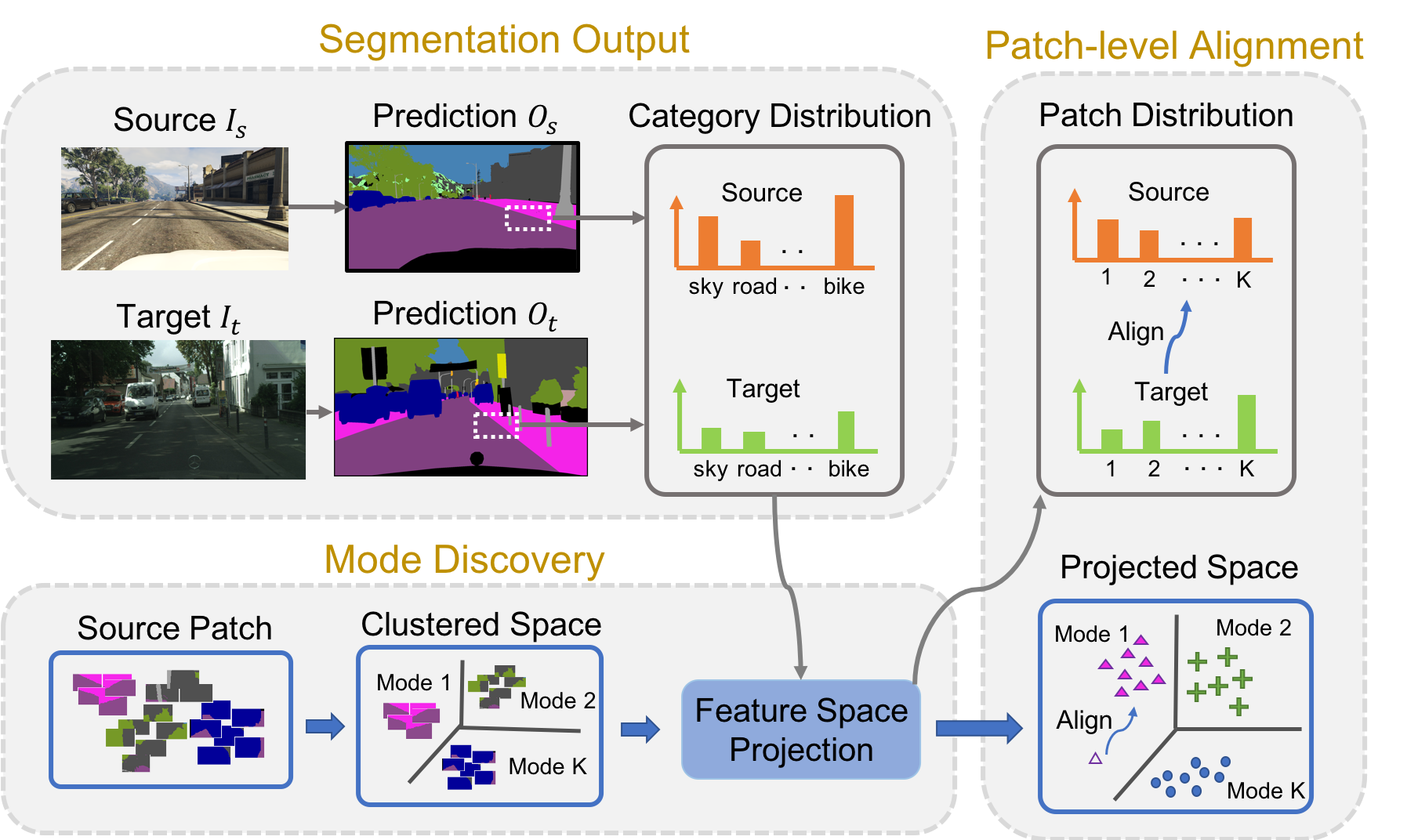}\\
	\vspace{1mm}
	\caption{
		An overview of our patch-level alignment.
		For our method, the category distribution is projected to the patch distribution via a clustered space that is constructed by discovering $K$ patch modes in the source domain.
		For the target data, we then align patch distributions across domains via adversarial learning in this $K$-dimensional space.
		Note that, compared to the output space adaptation methods, they only have a step that directly aligns category distributions without consider multiple modes in the source data.
		%
	}
	\label{fig: intro}
	\vspace{-3mm}
\end{figure*}
\section{Related Work}
%
We discuss unsupervised domain adaptation methods for image classification and pixel-level structured prediction tasks, and works on learning disentangled representations.
%
\paragraph{UDA for Image Classification.}
UDA methods have been developed for classification by aligning the feature distributions between the source and the target domains.
Conventional methods use hand-crafted features~\cite{Fernando_ICCV_2013,gong2012geodesic} to minimize the discrepancy across domains, while recent algorithms utilize deep architectures~\cite{ganin2016domain, tzeng2015simultaneous} to learn domain-invariant features.
One common practice is to adopt adversarial learning~\cite{ganin2016domain} or to minimize the Maximum Mean Discrepancy~\cite{long2015learning}.
Several variants have been developed by designing different classifiers~\cite{long2016unsupervised} and loss functions~\cite{tzeng2015simultaneous,tzeng2017adversarial}, and for distance metric learning~\cite{sohn2017unsupervised,sohn2019unsupervised}.
In addition, other recent work aims to enhance feature representations by pixel-level transfer~\cite{Bousmalis_CVPR_2017} and maximum classifier discrepancy~\cite{Saito_CVPR_2018}.

\paragraph{UDA for Semantic Segmentation.}
Following the practice in image classification, domain adaptation for pixel-level predictions has been studied. 
%
%
\cite{Hoffman_CoRR_2016} introduces to tackle the semantic segmentation problem for road-scene images by adapting from synthetic images via aligning global feature representations.
%
%
In addition, a category-specific prior, e.g., object size and class distribution is extracted from the source domain and is transferred to the target distribution as a constraint.
%
%
Instead of designing such constraints, \cite{Zhang_ICCV_2017} applies the SVM classifier to capture label distributions on superpixels as the property to train the adapted model.
Similarly, \cite{Chen_ICCV_2017} proposes a class-wise domain adversarial alignment by assigning pseudo labels to the target data.
%
%
%

More recently, numerous approaches are proposed to improve the adapted segmentation and can be categorized as follows:
1) output space~\cite{Tsai_CVPR_2018} and spatial-aware~\cite{Chen_CVPR_2018} adaptations aim to align the global structure (e.g., scene layout) across domains;
2) pixel-level adaptation synthesizes target samples~\cite{Hoffman_ICML_2018, Murez_CVPR_2018, Wu_ECCV_2018, Zhang_CVPR_2018} to reduce the domain gap during training the segmentation model;
3) pseudo label re-training~\cite{Saleh_ECCV_2018, Zou_ECCV_2018} generates pseudo ground truth of target images to finetune the model trained on the source domain.
While the most relevant approaches to ours are from the first category, they do not handle the intrinsic domain gap such as camera poses.
In contrast, the proposed patch-level alignment is able to match patches at various image locations across domains.
We also note that, the other two categories or other techniques such as robust loss function design~\cite{Zhu_ECCV_2018} are orthogonal to the contribution of this work.
In Section~\ref{sec:improve}, we show that our patch-level representations can be integrated with other domain adaptation methods to further enhance the performance.
%
%
%
\vspace{-3mm}
\paragraph{Learning Disentangled Representations.}
Learning a latent disentangled space has led to a better understanding for numerous tasks such as facial recognition~\cite{Reed_ICML_2014}, image generation~\cite{Chen_NIPS_2016, Odena_ICML_2017}, and view synthesis~\cite{Kulkarni_NIPS_2015, Yang_NIPS_2015}.
These approaches use predefined factors to learn interpretable representations of the image.
\cite{Kulkarni_NIPS_2015} propose to learn graphic codes that are disentangled with respect to various image transformations, e.g., pose and lighting, for rendering 3D images.
Similarly, \cite{Yang_NIPS_2015} synthesize 3D objects from a single image via an encoder-decoder architecture that learns latent representations based on the rotation factor.
Recently, AC-GAN~\cite{Odena_ICML_2017} develops a generative adversarial network (GAN) with an auxiliary classifier conditioned on the given factors such as image labels and attributes.

Although these methods present promising results on using the specified factors and learning a disentangled space to help the target task, they focus on handling the data in a single domain.
Motivated by this line of research, we propose to learn discriminative representations for patches to help the domain adaptation task.
To this end, we take advantage of the available label distributions and naturally utilize them as a disentangled factor, in which our framework does not require to predefine any factors like conventional methods.

%% file: method.tex
%
\begin{figure*}[t]
	\centering
	\includegraphics[width=0.85\linewidth]{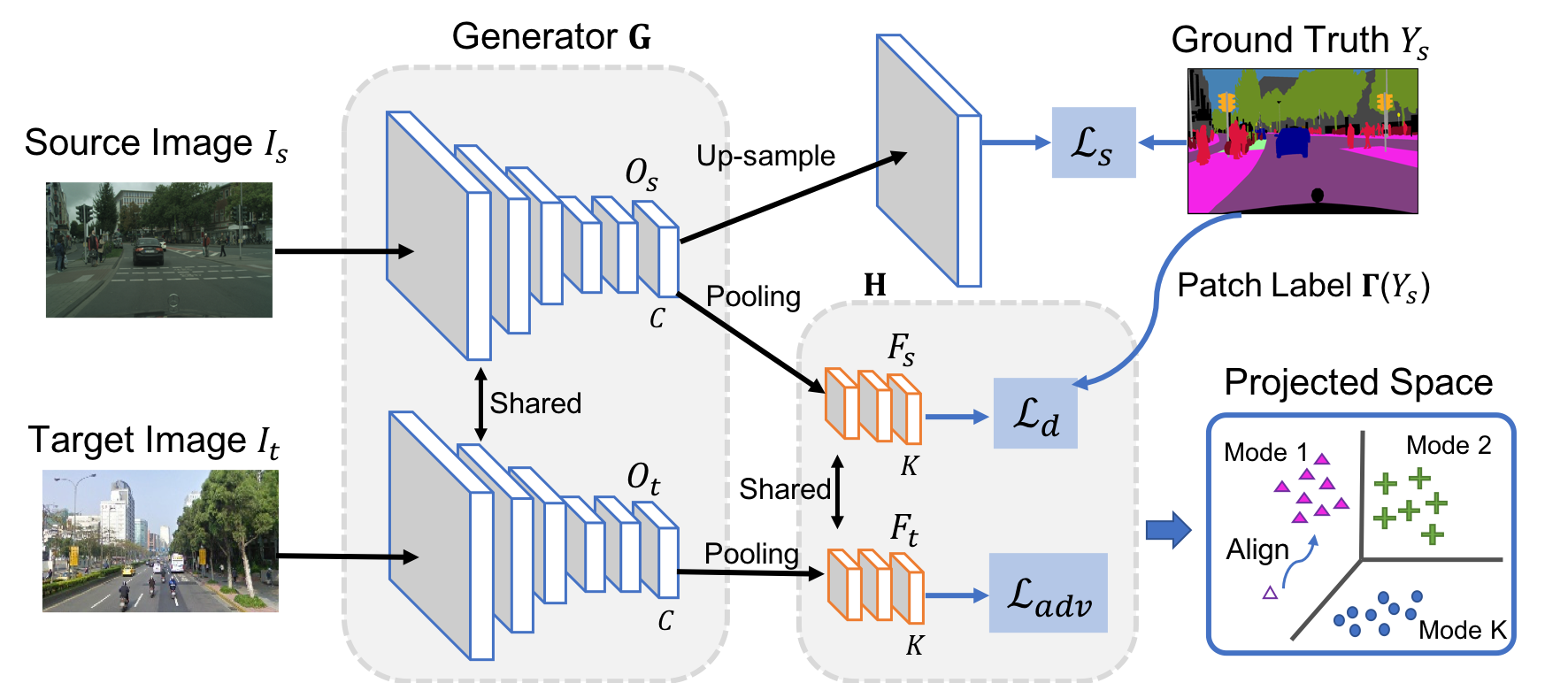} \\
	\vspace{1mm}
	\caption{The proposed network architecture that consists of a generator $\mathbf{G}$ and a categorization module $\mathbf{H}$ for learning discriminative patch representations via 1) patch mode discovery supervised by the patch classification loss $\mathcal{L}_{d}$, and 2) patch-level alignment via the adversarial loss $\mathcal{L}_{adv}$.
		In the projected space, solid symbols denote source representations and unfilled ones are target representations pulled to the source distribution.
	}
	\label{fig: architecture}
		\vspace{-3mm}
\end{figure*}
\section{Domain Adaptation for Structured Output}
In this section, we describe our framework for predicting structured outputs: an adversarial learning scheme to align distributions across domains by using discriminative output representations of patches.
\subsection{Algorithm Overview}
%
%
Given the source and target images $I_s, I_t \in \mathbb{R}^{H \times W \times 3}$, where only the source data is annotated with per-pixel semantic categories $Y_s$, we seek to learn a semantic segmentation model $\mathbf{G}$ that works on both domains.
Since the target domain is unlabeled, our goal is to align the predicted output distribution $O_t$ of the target data with the source distribution $O_s$, which is similar to~\cite{Tsai_CVPR_2018}.
%
%
However, such distribution is not aware of the local difference in patches and thus is not able to discover a diverse set of modes during adversarial learning.
%
To tackle this issue, and in contrast to~\cite{Tsai_CVPR_2018}, we project the category distribution of patches to the clustered space that already discovers various patch modes (\ie, $K$ clusters) based on the annotations in the source domain.
%
%
For target data, we then employ adversarial learning to align the patch-level distributions across domains in the $K$-dimensional space.
%
\subsection{Patch-level Alignment}
As in Figure~\ref{fig: intro}, we seek for ways to align patches in a clustered space that provides a diverse set of patch modes.
One can also treat this procedure as learning prototypical output representations of patches by clustering ground truth segmentation annotations from the source domain.
In what follows, we introduce how we construct the clustered space and learn discriminative patch representations. Then we describe adversarial alignment using the learned patch representation.
The detailed architecture is shown in Figure~\ref{fig: architecture}.
\paragraph{Patch Mode Discovery.}
To discover modes and learn a discriminative feature space, class labels~\cite{salimans2016improved} or predefined factors~\cite{Odena_ICML_2017} are usually provided as supervisory signals.
However, it is non-trivial to assign a class membership to individual patches of an image. One may apply unsupervised clustering of image patches, but it is unclear whether the constructed clustering would separate patches in a semantically meaningful way.
In this work, we make use of per-pixel annotations available in the source domain to construct a space of semantic patch representation.
To achieve this, we use label histograms for patches.
We first randomly sample patches from source images, use a 2-by-2 grid on patches to extract spatial label histograms, and concatenate them to obtain a $2\,{\times}\,2\,{\times}\,C$ dimensional vector.
%
Second, we apply K-means clustering on these histograms, thereby assigning each ground truth label patch a unique cluster index.
We define the process of finding the cluster membership for each patch in a ground truth label map $Y_s$ as $\Gamma(Y_s)$.
%
%

To incorporate this clustered space for training the segmentation network $\mathbf{G}$ on source data, we add a classification module $\mathbf{H}$ on top of the predicted output $O_s$, which tries to predict the cluster membership $\Gamma(Y_s)$ for all locations.
We denote the learned representation as $F_s = \mathbf{H}(\mathbf{G}(I_s)) \in(0,1)^{U \times V \times K}$ through the softmax function, where $K$ is the number of clusters.
%
Here, each data point on the spatial map $F_s$ corresponds to a patch of the input image, and we obtain the group label for each patch via $\Gamma(Y_s)$.
Then the learning process to construct the clustered space can be formulated as a cross-entropy loss:
\begin{equation}
	\mathcal{L}_{d}(F_s, \Gamma(Y_s); \mathbf{G}, \mathbf{H}) =
	- \sum_{u,v}{} \sum_{k \in K} CE^{(u,v,k)} \;,
	\label{eq:loss_dis}
\end{equation}
where $CE^{(u,v,k)} = \Gamma(Y_s)^{(u,v,k)} \log(F_s^{(u,v,k)})$.
\vspace{-3mm}
\paragraph{Adversarial Alignment.}
The ensuing task is to align the representations of target patches to the clustered space constructed in the source domain, ideally aligned to one of the $K$ modes.
To this end, we utilize an adversarial loss between $F_s$ and $F_t$, where $F_t$ is generated in the same way as described above.
Note that, the patch-level feature $F$ is now transformed from the category distribution $O$ to the clustered space defined by $K$-dimensional vectors.
%
%
%
We then formulate the patch distribution alignment in an adversarial objective:
\begin{align}
	\label{eq:loss_local}
	\mathcal{L}_{adv}(F_s, F_t; \mathbf{G}, \mathbf{H}, \mathbf{D}) = \sum_{u,v}{} \mathbb{E} [\log \mathbf{D}(F_s)^{(u,v,1)}] \\ \notag
	+ \mathbb{E} [\log (1- \mathbf{D}(F_t)^{(u,v,1)})],
\end{align}
where $\mathbf{D}$ is the discriminator to classify whether the feature representation $F$ is from the source or the target domain.
%
%


\vspace{-3mm}
\paragraph{Learning Objective.}
%
We integrate \eqref{eq:loss_dis} and \eqref{eq:loss_local} into the min-max problem (for clarity, we drop all arguments to losses except the optimization variables):
\begin{align}\
\label{eq:optimize_all}
\min_{\mathbf{G}, \mathbf{H}} \ \max_{\mathbf{D}} \mathcal{L}_{s}(\mathbf{G}) +  \lambda_d\mathcal{L}_{d}(\mathbf{G}, \mathbf{H}) \\ \notag
+ \lambda_{adv}\mathcal{L}_{adv}(\mathbf{G}, \mathbf{H}, \mathbf{D}),
\end{align}
%
%
%
%
where $\mathcal{L}_{s}$ is the supervised cross-entropy loss for learning the structured prediction (e.g., semantic segmentation) on source data, and $\lambda$'s are the weights for different losses.
\subsection{Network Optimization}
%
To solve the optimization problem in Eq.~\eqref{eq:optimize_all}, we follow the procedure of training GANs~\cite{Goodfellow_NIPS_2014} and alternate two steps:
1) update the discriminator $\mathbf{D}$, and 2) update the networks $\mathbf{G}$ and $\mathbf{H}$ while fixing the discriminator.
%
%
\vspace{-3mm}
\paragraph{Update the Discriminator $\mathbf{D}$.}
We train the discriminator $\mathbf{D}$ to classify whether the feature representation $F$ is from the source (labeled as 1) or the target domain (labeled as 0).
The maximization problem wrt. $\mathbf{D}$ in \eqref{eq:optimize_all} is equivalent to minimizing the binary cross-entropy loss:
%
%
\begin{align}
\mathcal{L}_{\mathbf{D}}(F_s, F_t; \mathbf{D}) = - \sum_{u,v}{} \log(\mathbf{D}(F_s)^{(u,v,1)}) \\ \notag
+ \log(1-\mathbf{D}(F_t)^{(u,v,1)}).
\label{eq:loss_d_l}
\end{align}
\vspace{-3mm}
\paragraph{Update the Networks $\mathbf{G}$ and $\mathbf{H}$.}
The goal of this step is to push the target distribution closer to the source distribution using the optimized $\mathbf{D}$, while maintaining good performance on the main tasks using $\mathbf{G}$ and $\mathbf{H}$.
As a result, the minimization problem in \eqref{eq:optimize_all} is the combination of two supervised loss functions with the adversarial loss, which can be expressed as a binary cross-entropy function that assigns the source label to the target distribution:
\begin{align}
\mathcal{L}_{\mathbf{G,H}} = \mathcal{L}_{s} + \lambda_d\mathcal{L}_{d} - \lambda_{adv} \sum_{u,v}{} \log(\mathbf{D}(F_t)^{(u,v,1)}).
\end{align}
%
%
We note that updating $\mathbf{H}$ also influences $\mathbf{G}$ through back-propagation, and thus the feature representations are enhanced in $\mathbf{G}$.
%
In addition, we only require $\mathbf{H}$ during the training phase, so that runtime for inference is unaffected compared to the output space adaptation approach~\cite{Tsai_CVPR_2018}.
\subsection{Implementation Details}
%
%
\paragraph{Network Architectures.}
The generator consists of the network $\mathbf{G}$ with a categorization module $\mathbf{H}$.
For a fair comparison, we follow the framework used in~\cite{Tsai_CVPR_2018} that adopts DeepLab-v2~\cite{deeplab} with the ResNet-101 architecture~\cite{He_CVPR_2016} as our baseline network $\mathbf{G}$.
To add the module $\mathbf{H}$ on the output prediction $O$, we first use an adaptive average pooling layer to generate a spatial map, where each data point on the map has a desired receptive field corresponding to the size of extracted patches.
Then this pooled map is fed into two convolution layers and a feature map $F$ is produced with the channel number $K$. Figure~\ref{fig: architecture} illustrates the main components of the proposed architecture.
For the discriminator $\mathbf{D}$, input data is a $K$-dimensional vector and we utilize 3 fully-connected layers similar to~\cite{tzeng2017adversarial}, with leaky ReLU activation and channel numbers $\{256, 512, 1\}$.
\vspace{-3mm}
\paragraph{Implementation Details.}
We implement the proposed framework using the PyTorch toolbox on a single Titan X GPU with 12 GB memory.
To train the discriminators, we use the Adam optimizer~\cite{Kingma_ICLR_2015} with initial learning rate of $10^{-4}$ and momentums set as 0.9 and 0.99.
For learning the generator, we use the Stochastic Gradient Descent (SGD) solver where the momentum is 0.9, the weight decay is $5\times10^{-4}$ and the initial learning rate is $2.5\times10^{-4}$.
For all the networks, we decrease the learning rates using the polynomial decay with a power of 0.9, as described in~\cite{deeplab}.
During training, we select $\lambda_d = 0.01$, $\lambda_{adv} = 0.0005$ and $K = 50$ fixed for all the experiments.
Note that we first train the model only using the loss $\mathcal{L}_{s}$ for 10K iterations to avoid initially noisy predictions and then train the network using all the loss functions.
More details of the hyper-parameters such as image and patch sizes are provided in the supplementary material.

%% file: experiment.tex
\begin{figure*}[t!]
	\centering
	\includegraphics[width=0.9\linewidth]{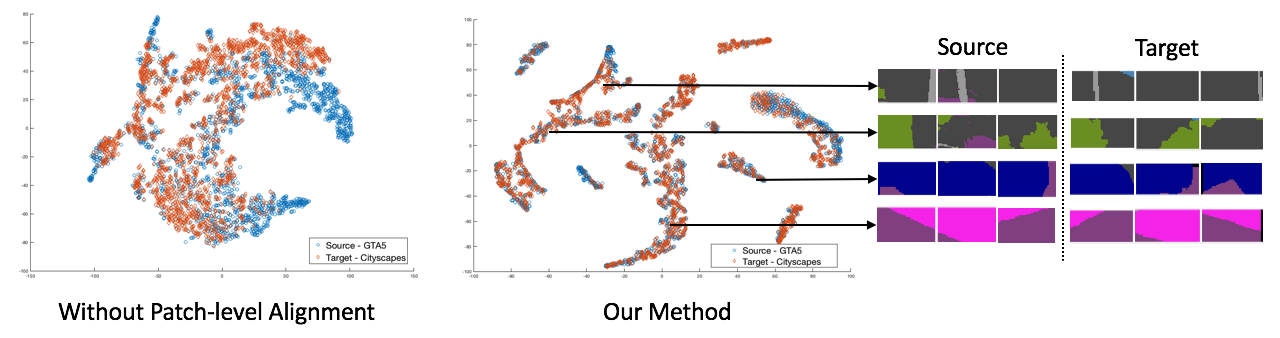}
	\caption{Visualization of patch-level representations. We first show feature representations via t-SNE of our method and compare to the baseline without the proposed patch-level alignment. In addition, we show patch examples in the clustered space. In each group, patches are similar in appearance (i.e., each color represents a semantic label) between the source and target domains.
	}
	\vspace{-1mm}
	\label{fig: feature}
\end{figure*}
		
		
		
		
%
\section{Experimental Results}
We evaluate the proposed framework for domain adaptation on semantic segmentation.
We first conduct an extensive ablation study to validate key components of our algorithm.
Second, we show that the proposed method can be integrated with various domain adaptation techniques, including output space adaptation~\cite{Tsai_CVPR_2018}, pixel-level adaptation~\cite{Hoffman_ICML_2018}, and pseudo label re-training~\cite{Zou_ECCV_2018}.
This demonstrates that our learned patch-level representations are complementary to a wide range of domain adaptation strategies and provide additional benefits.
Finally, we present a hybrid model that performs favorably against state-of-the-art approaches on numerous benchmark datasets and settings.
\subsection{Evaluated Datasets and Metric}
We evaluate our domain adaptation method on semantic segmentation under various settings, including synthetic-to-real and cross-city.
First, we adapt the synthetic GTA5~\cite{Richter_ECCV_2016} dataset to the Cityscapes~\cite{cityscapes} dataset that contains real road-scene images.
Similarly, we use the SYNTHIA~\cite{Ros_CVPR_2016} dataset, which has a larger domain gap to Cityscapes images.
For these experiments, we follow~\cite{Hoffman_CoRR_2016} to split data into training and test sets.
As another example with high practical impact, we apply our method on data captured in different cities and weather conditions by adapting Cityscapes with sunny images to the Oxford RobotCar~\cite{oxford_robot} dataset containing rainy scenes.
%
We manually select 10 sequences in the Oxford RobotCar dataset tagged as ``rainy'' and randomly split them into 7 sequences for training and 3 for testing.
We sequentially sample 895 images for training and annotate 271 images with per-pixel semantic segmentation ground truth as the test set for evaluation.
The annotated ground truths are made publicly available at the project page.
For all experiments, Intersection-over-Union (IoU) ratio is used as the evaluation metric.
%
%
%
%
%
%
%
%
\subsection{Ablation Study and Analysis}
\begin{figure}[t!]
\vspace{-3mm}
	\centering
	\includegraphics[width=0.8\linewidth,trim={0 0 0 2cm},clip]{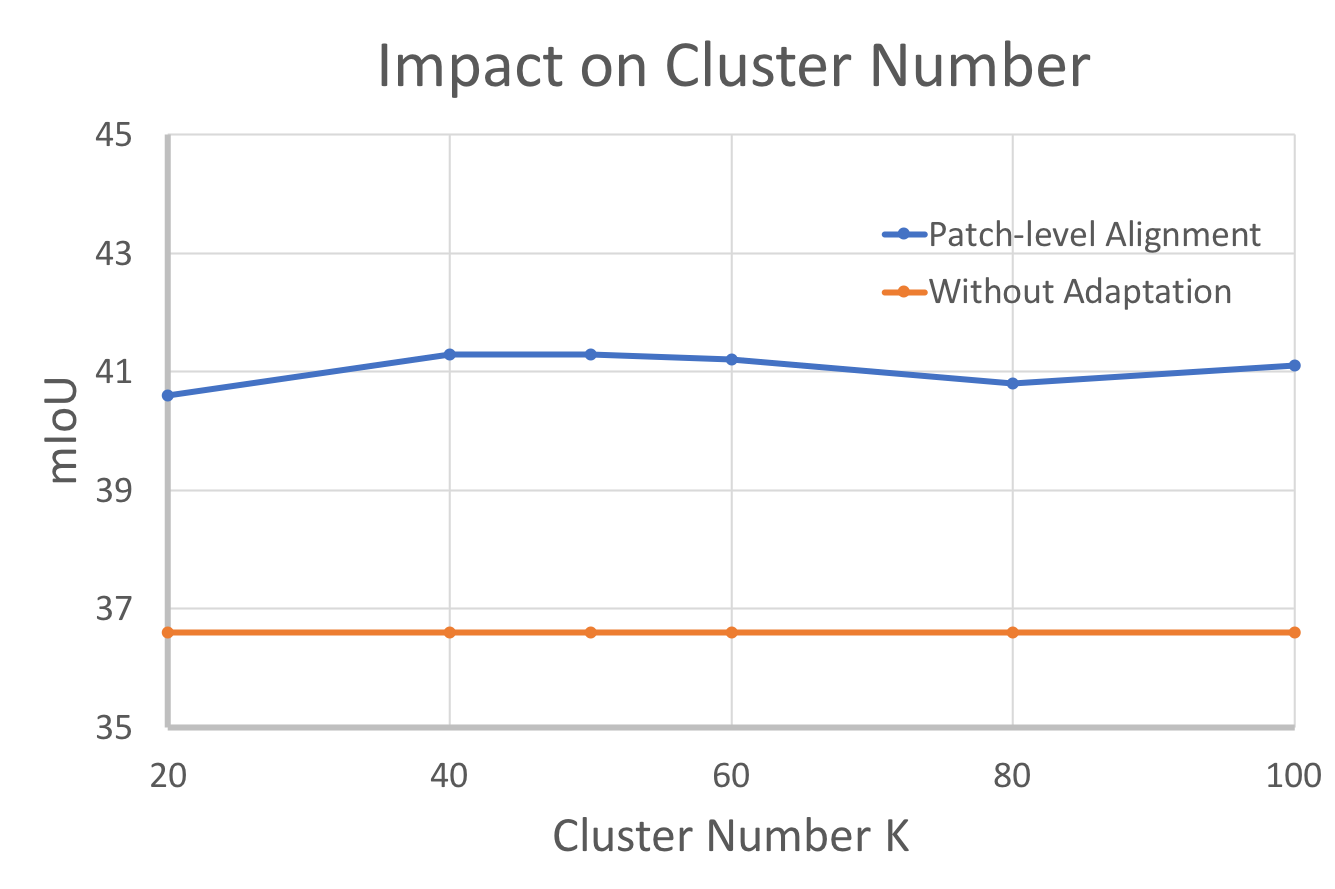}
	\vspace{-1mm}
	\caption{
	The performance of our method with respect to different numbers of clusters $K$ on GTA5-to-Cityscapes.
	}
	\label{fig: cluster}
\end{figure}

In Table~\ref{table:ablation_study}, we conduct the ablation study and analysis of the proposed patch-level alignment on the GTA5-to-Cityscapes scenario to understand the impact of different loss functions and design choices in our framework.
%
%
\begin{table} [t]
	\caption{
		Ablation study of the proposed loss functions on GTA5-to-Cityscapes using the ResNet-101 network.
	}
	\vspace{2mm}
	\label{table:ablation_study}
	\small
	\centering
	\renewcommand{\arraystretch}{1.1}
	\setlength{\tabcolsep}{6pt}
	\begin{tabular}{llc}
		\toprule
		
		\multicolumn{3}{c}{GTA5 $\rightarrow$ Cityscapes} \\
		\midrule
		
		Method & Loss Func. & mIoU \\
		
		\midrule
		
		Without Adaptation & $\mathcal{L}_{s}$ & 36.6 \\
		
		Discriminative Feature & $\mathcal{L}_{s} + \mathcal{L}_{d}$ & 38.8 \\
		
		Patch-level Alignment & $\mathcal{L}_{s} + \mathcal{L}_{d} + \mathcal{L}_{adv}$ & 41.3 \\
		\bottomrule
	\end{tabular}
\end{table}
\vspace{-4mm}
\paragraph{Loss Functions.}
In Table~\ref{table:ablation_study}, we show different steps of the proposed method, including the model without adaptation, using discriminative patch features and the final patch-level alignment.
Interestingly, we find that adding discriminative patch representations without any alignments ($\mathcal{L}_{s} + \mathcal{L}_{d}$) already improves the performance (from 36.6\% to 38.8\%), which demonstrates that the learned feature representation enhances the discrimination and generalization ability.
Finally, the proposed patch-level adversarial alignment improves the mIoU by 4.7\%.
%
%
%
\vspace{-3mm}
\paragraph{Impact of Learning Clustered Space.}
$K$-means provides an additional signal to separate different patch patterns, while performing alignment in this clustered space. Without the clustered loss $L_d$, it would be difficult to align patch modes across two domains. To validate it, we run an experiment by only using $L_s$ and $L_{adv}$ but removing $L_d$, and the performance is reduced by $1.9\%$ compared to our method ($41.3\%$). This shows the importance of learning the clustered space supervised by the $K$-means process.
\vspace{-3mm}
\paragraph{Impact of Cluster Number $K$.}
In Figure~\ref{fig: cluster}, we study the impact of the cluster number $K$ used to construct the patch representation, showing that the performance is robust to $K$. However, when $K$ is too large, e.g., larger than 300, it would cause confusion between patch modes and increases the training difficulty. To keep both efficiency and accuracy, we use $K=50$ throughout the experiments.
\vspace{-3mm}
\paragraph{Visualization of Feature Representations.}
In Figure~\ref{fig: feature}, we show the t-SNE visualization~\cite{maaten2008visualizing} of the patch-level features in the clustered space of our method and compare with the one without patch-level adaptation.
The result shows that with adaptation in the clustered space, the features are embedded into groups and the source/target representations overlap well.
In addition, we present example source/target patches with high similarity.
\begin{figure*}[t]
	\centering
	\begin{tabular}
		{@{\hspace{0mm}}c@{\hspace{1mm}} @{\hspace{0mm}}c@{\hspace{1mm}} @{\hspace{0mm}}c@{\hspace{1mm}}
			@{\hspace{0mm}}c@{\hspace{1mm}} @{\hspace{0mm}}c@{\hspace{0mm}}
		}
		
		\includegraphics[width=0.18\linewidth]{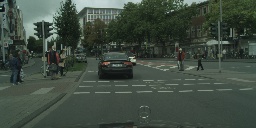} &
		\includegraphics[width=0.18\linewidth]{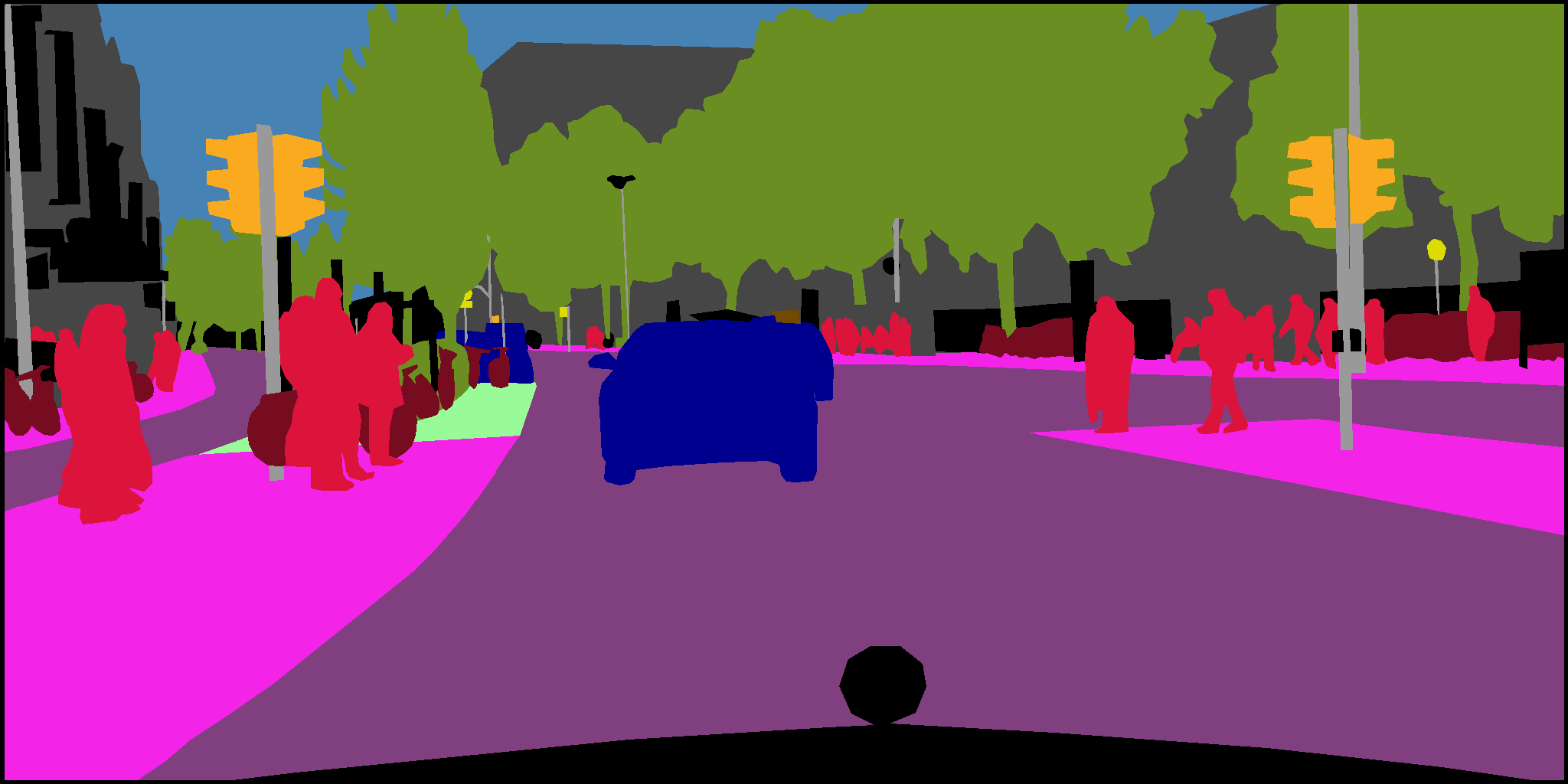} &
		\includegraphics[width=0.18\linewidth]{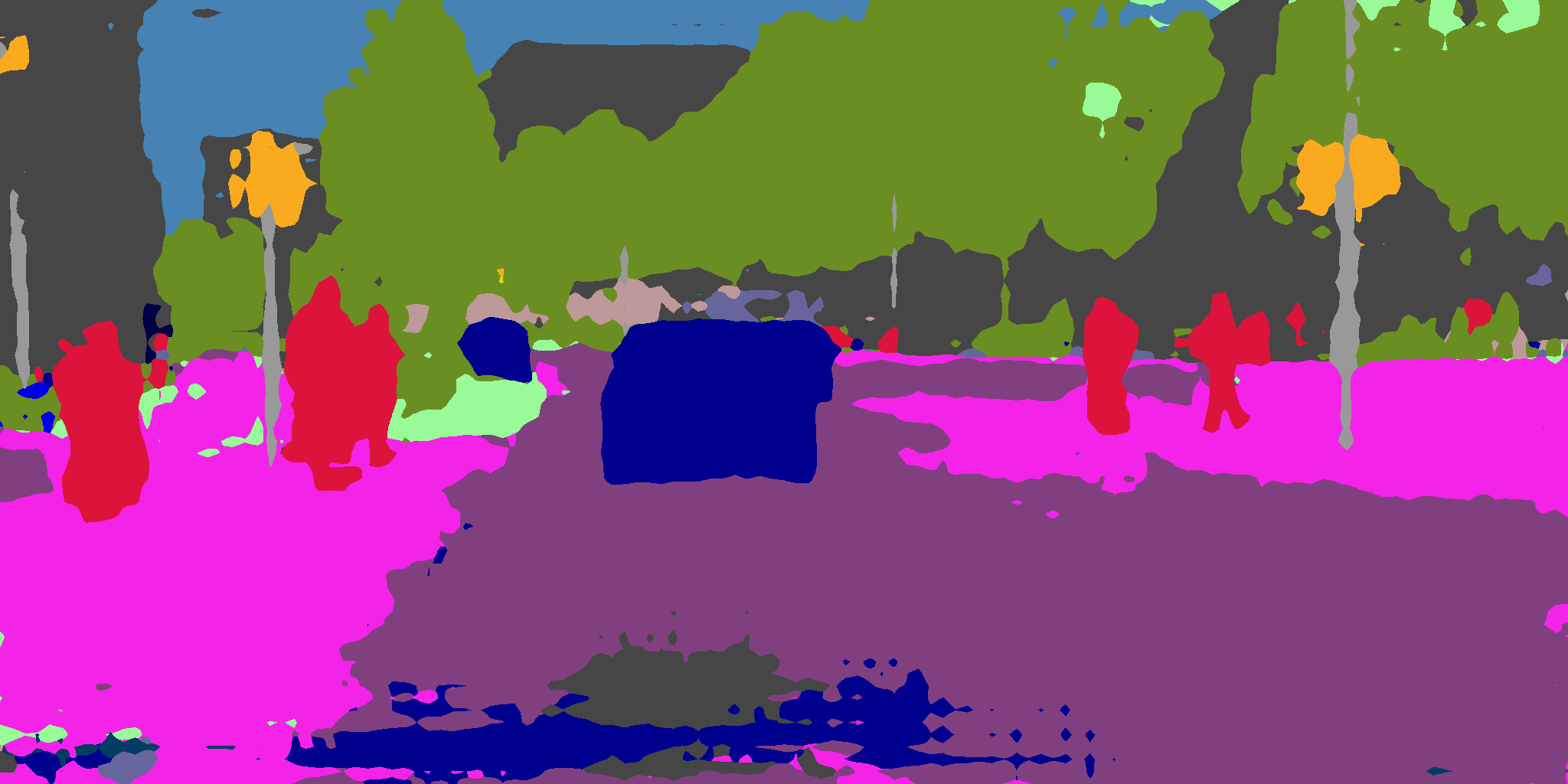} &
		\includegraphics[width=0.18\linewidth]{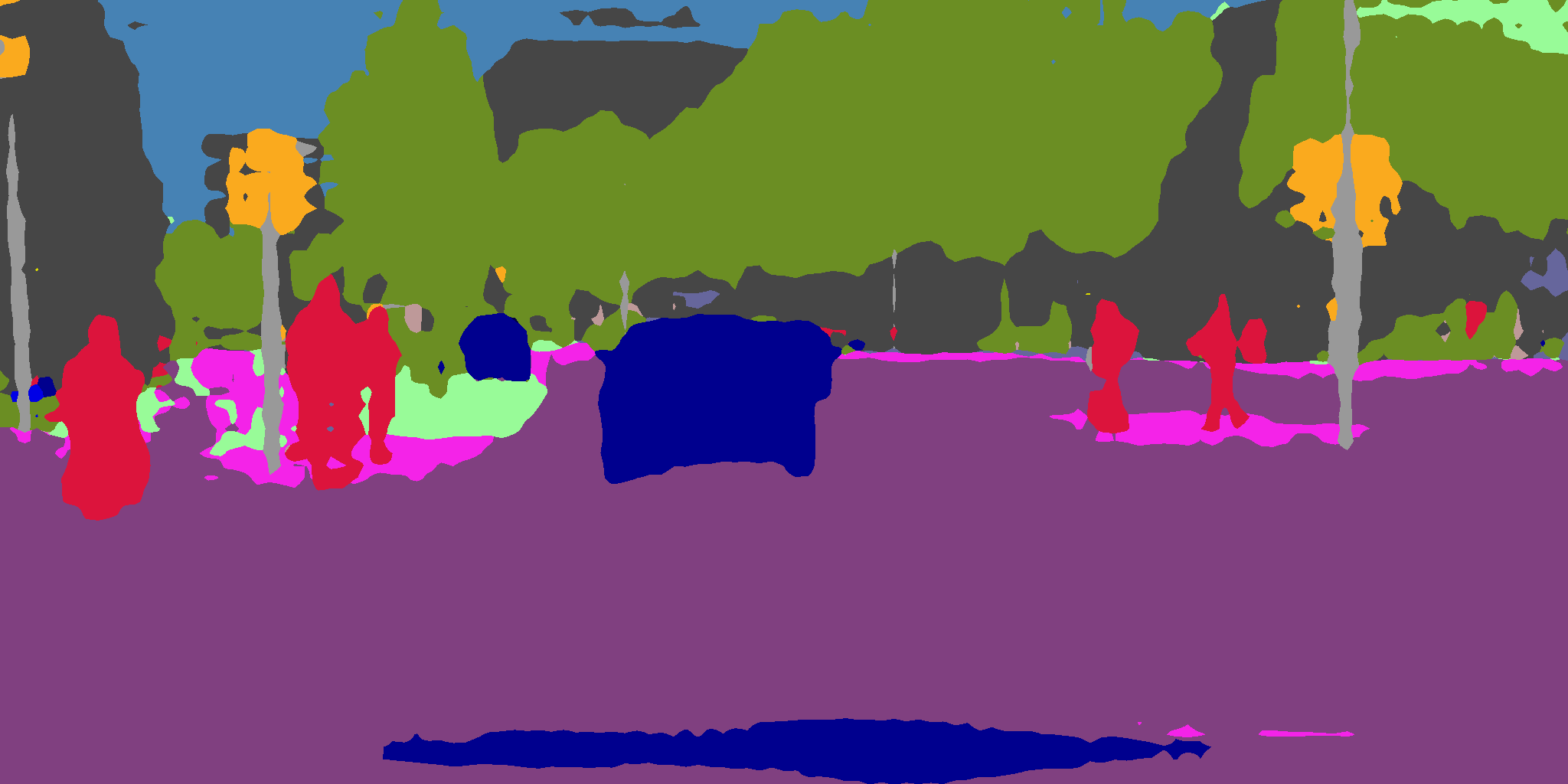} &
		\includegraphics[width=0.18\linewidth]{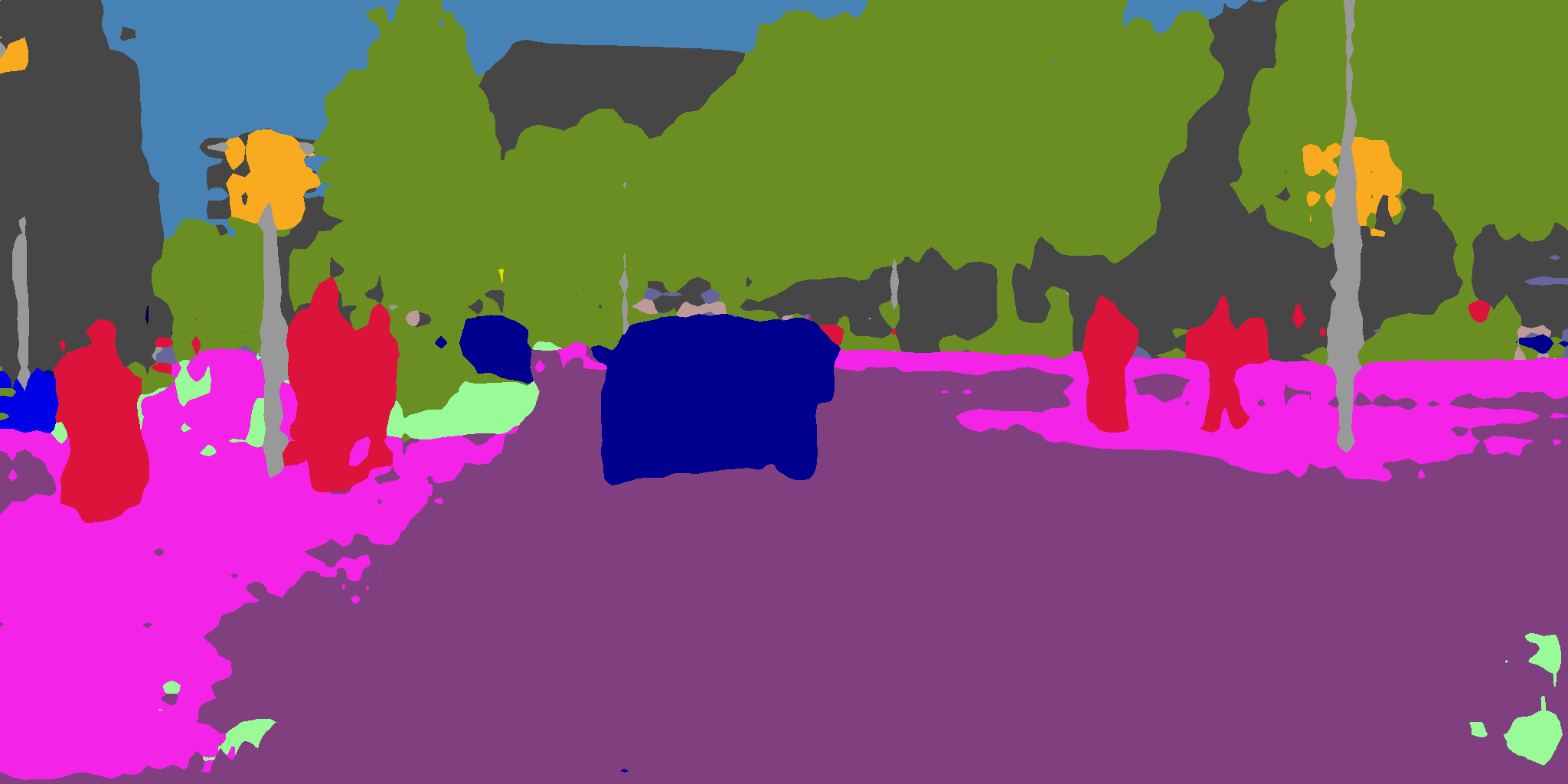} \\
		
		\includegraphics[width=0.18\linewidth]{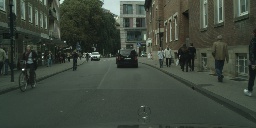} &
		\includegraphics[width=0.18\linewidth]{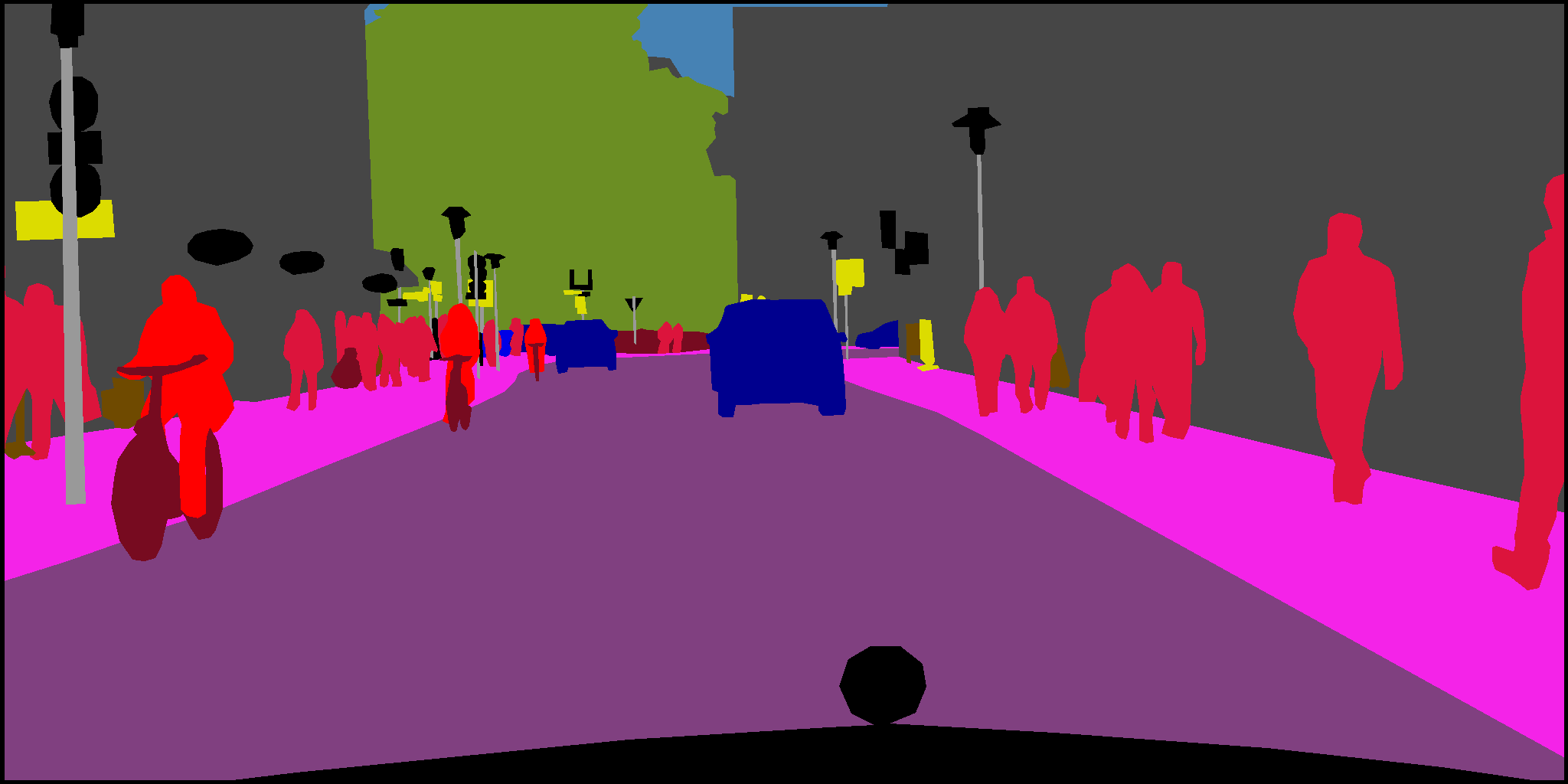} &
		\includegraphics[width=0.18\linewidth]{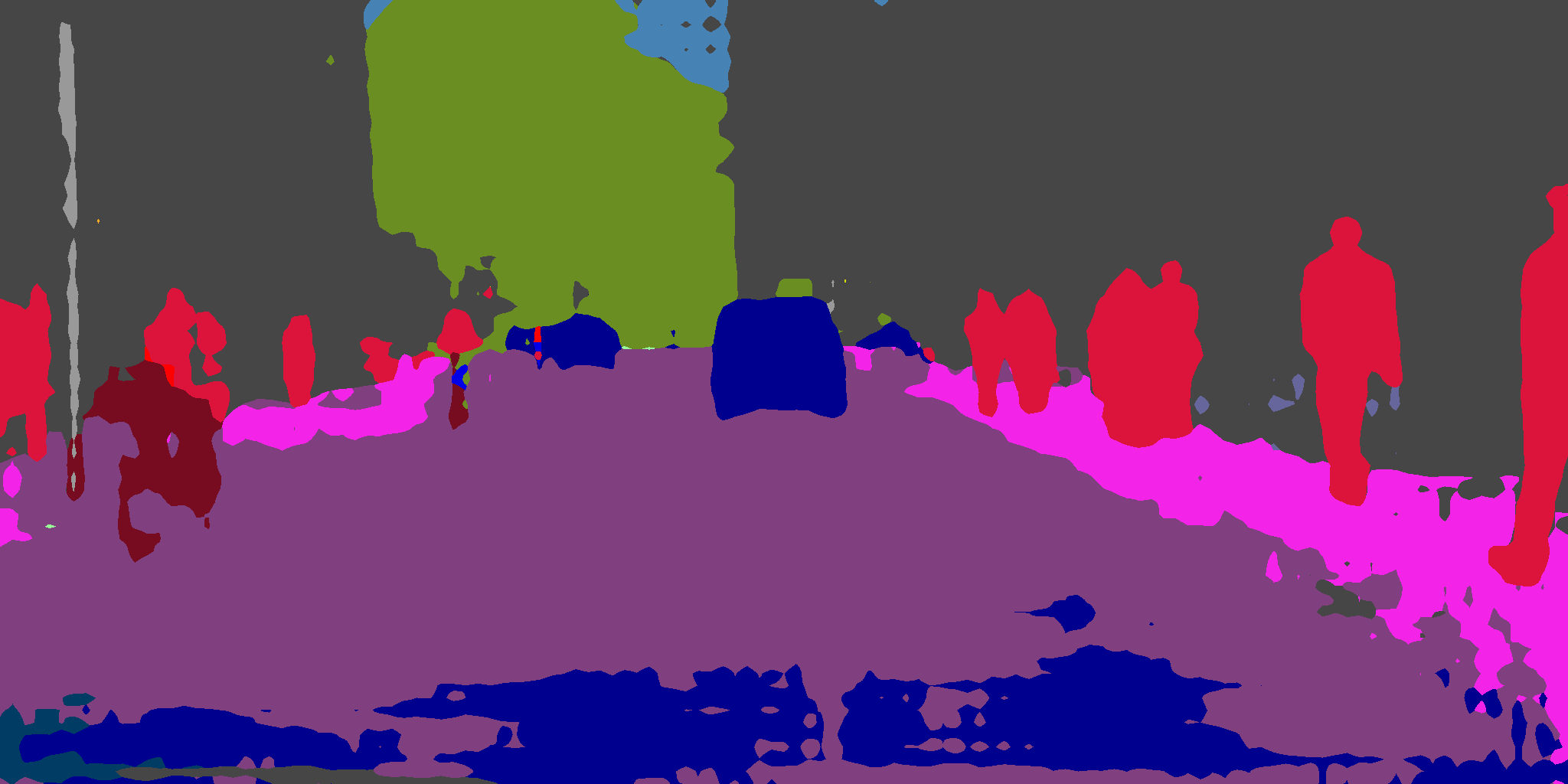} &
		\includegraphics[width=0.18\linewidth]{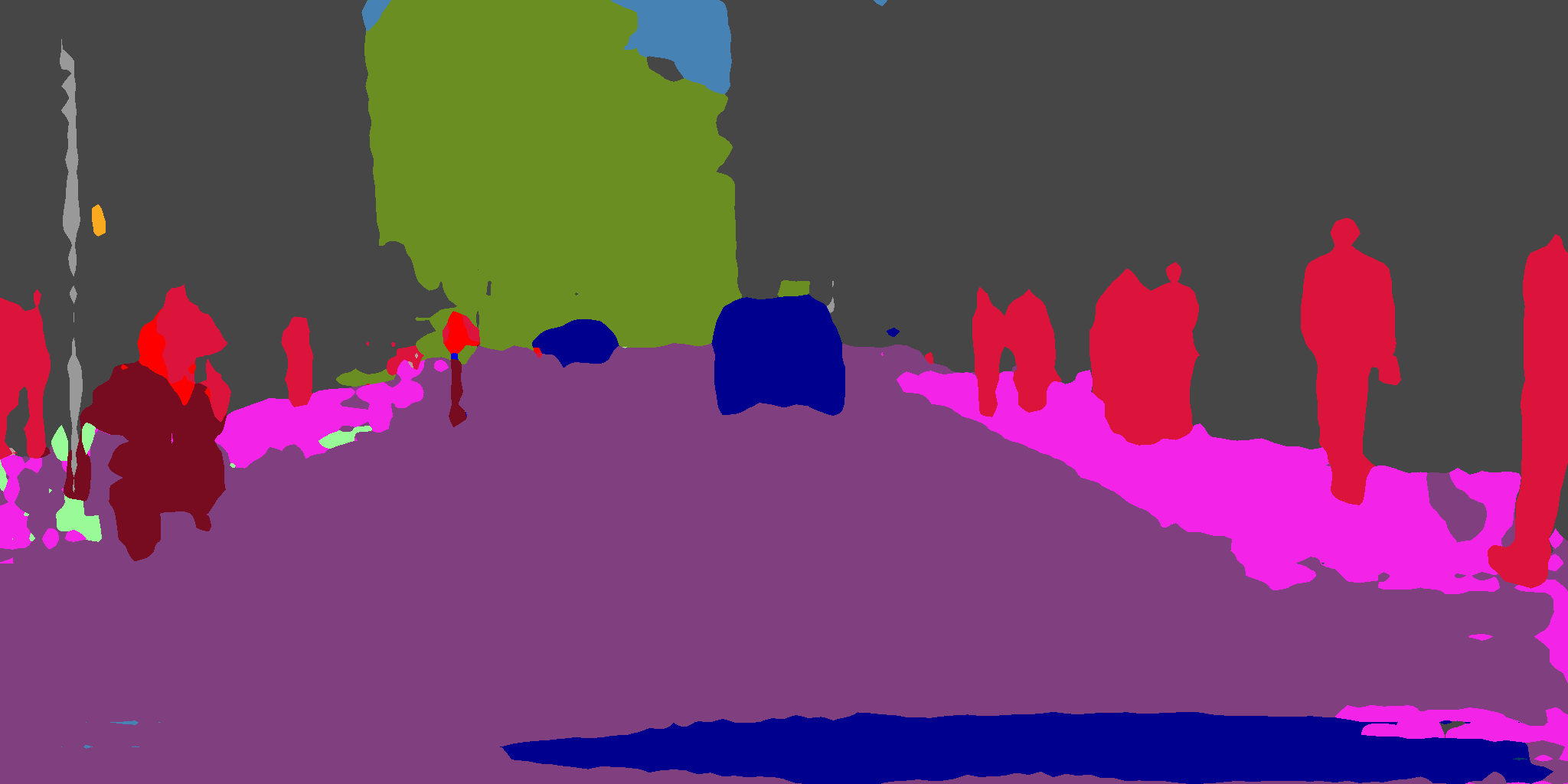} &
		\includegraphics[width=0.18\linewidth]{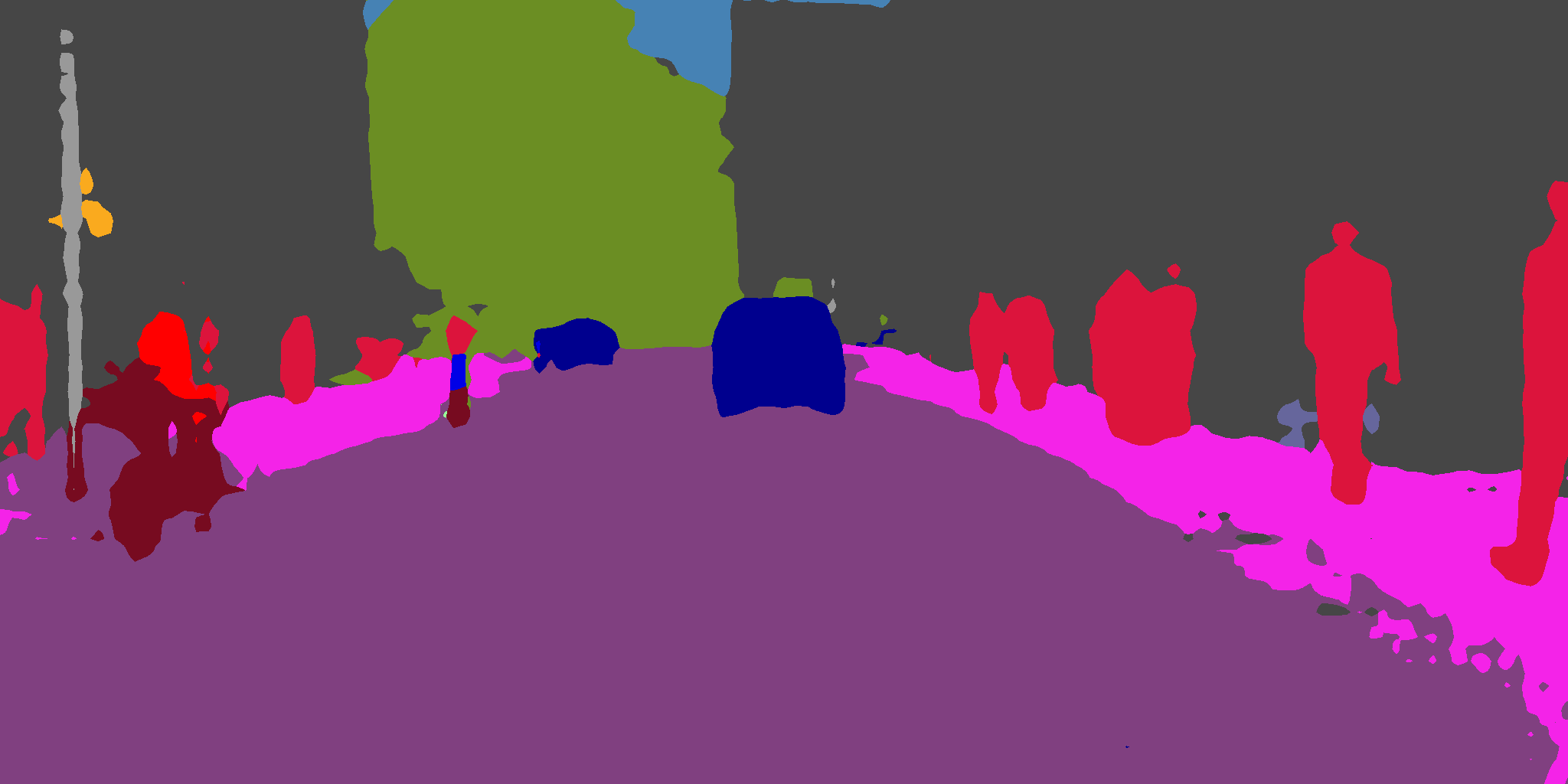} \\
		
		Target Image & Ground Truth & Before Adaptation & Output Alignment & Patch Alignment \\
		
	\end{tabular}
	\caption{
		Example results for GTA5-to-Cityscapes. Our method often generates the segmentation with more details (e.g., sidewalk and pole) while producing less noisy regions compared to the output space adaptation approach~\cite{Tsai_CVPR_2018}.
	}
	\vspace{-3mm}
	\label{fig: visual}
\end{figure*}
\subsection{Improvement on Domain Adaptation Methods}
\label{sec:improve}
%
The learned patch representation via the proposed patch alignment enhances feature representations and is complementary to various DA methods, which we demonstrate by combining with output-space adaptation (Ou), pixel-level adaptation (Pi) and pseudo label re-training (Ps).  Our results show consistent improvement in all cases, \eg, 1.8\% to 2.7\% on GTA5-to-Cityscapes, as shown in Table~\ref{table:improve}.
\vspace{-3mm}
\paragraph{Output Space Adaptation.}
%
We first consider methods that align the {\em global} layout across domains as in \cite{Chen_CVPR_2018,Tsai_CVPR_2018}.
Our proposed cluster prediction network $\mathbf{H}$ and the corresponding loss $\mathcal{L}_{adv}$ can be simply added into~\cite{Tsai_CVPR_2018}.
Since these methods only align the global structure, adding our method helps figuring out local details better and improves the segmentation quality.
\vspace{-3mm}
\paragraph{Pixel-level Adaptation.}
We utilize CyCADA~\cite{Hoffman_ICML_2018} as the pixel-level adaptation algorithm and produce synthesized images in the target domain from source images.
To train our model, we add synthesized samples into the labeled training set with the proposed patch-level alignment.
Note that, since synthesized samples share the same pixel-level annotations as the source data, they can be also considered in our clustering process and the optimization in \eqref{eq:optimize_all}.
%
%
%
\vspace{-3mm}
\paragraph{Pseudo Label Re-training.}
Pseudo label re-training is a natural way to improve the segmentation quality in domain adaptation~\cite{Zou_ECCV_2018} or semi-supervised learning~\cite{Hung_BMVC_2018}.
%
The end-to-end trainable framework~\cite{Hung_BMVC_2018} uses an adversarial scheme to identify self-learnable regions, which makes it an ideal candidate to integrate our patch-level adversarial loss.
%
%
%
\begin{table} [t]
	\caption{
		Performance improvements in mIoU of integrating our patch-level alignment with existing domain adaptation approaches on GTA5-to-Cityscapes using the ResNet-101 network.
	}
	\vspace{1mm}
	\label{table:improve}
	\small
	\centering
	\renewcommand{\arraystretch}{1.1}
	\setlength{\tabcolsep}{6pt}
	\begin{tabular}{lccc}
		\toprule
		
		\multicolumn{4}{c}{GTA5 $\rightarrow$ Cityscapes (19 Categories)} \\
		\midrule
		
		Methods & Base & + Patch-Alignment & $\Delta$ \\
		
		\midrule
		Without Adaptation & 36.6 & 41.3 & +4.7 \\
		
		(Ou)tput Space Ada. & 41.4 & 43.2 & +1.8 \\

		(Pi)xel-level Ada. & 42.2 & 44.9 & +2.7 \\

		(Ps)eudo-GT & 41.8 & 44.2 & +2.4\\
		
		(Fu)sion & 44.5 & 46.5 & +2.0 \\
		
		\bottomrule
	\end{tabular}
\end{table}
\vspace{-3mm}
\paragraph{Results and Discussions.}
%
%
The results for combining the proposed patch-level alignment with the three above mentioned DA methods are shown in Tables~\ref{table:improve} and \ref{table:improve_synthia} for GTA5-to-Cityscapes and SYNTHIA-to-Cityscapes, respectively.
We can observe that adding patch-level alignment improves in all cases.
For reference, we also show the gain from adding patch-level alignment to the plain segmentation network (without adaptation).
Even when combining all three DA methods, i.e., Fusion (Fu), the proposed patch-alignment further improves the results significantly ($\ge 2.0\%$).
Note that, the combination of all DA methods including patch alignment, \ie, Fu + Patch-Alignment, achieves the best performance in both cases.

As a comparison point, we also try to combine pixel-level adaptation with output space alignment (Pi + Ou), but the performance is 0.7\% worse than ours, \ie, Pi + Patch-Alignment, showing the advantages of adopting patch-level alignment.
On SYNTHIA-to-Cityscapes in Table~\ref{table:improve_synthia}, we find that Pi and Ps are less effective than Ou, likely due to the poor quality of the input data in the source domain, which also explains the lower performance of the combined model (Fu).
This also indicates that directly combining different DA methods may not improve the performance incrementally. 
However, adding the proposed patch-alignment improves the results consistently in all settings.
\begin{table} [t]
	\caption{
		Performance improvements in mIoU of integrating our patch-level alignment with existing domain adaptation approaches on SYNTHIA-to-Cityscapes using the ResNet-101 network.
	}
	\vspace{1mm}
	\label{table:improve_synthia}
	\small
	\centering
	\renewcommand{\arraystretch}{1.1}
	\setlength{\tabcolsep}{6pt}
	\begin{tabular}{lccc}
		\toprule
		
		\multicolumn{4}{c}{SYNTHIA $\rightarrow$ Cityscapes (16 Categories)} \\
		\midrule
		
		Methods & Base & + Patch-Alignment & $\Delta$ \\
		
		\midrule
		Without Adaptation & 33.5 & 37.0 & +3.5 \\
		
		(Ou)tput Space Ada. & 39.5 & 39.9 & +0.4 \\

		(Pi)xel-level Ada. & 35.8 & 37.0 & +1.2\\

		(Ps)eudo-GT & 37.4 & 38.9 & +1.5 \\
		
		(Fu)sion & 37.9 & 40.0 & +2.1 \\
		
		\bottomrule
	\end{tabular}
\end{table}
\begin{table*} [t]
	\caption{Results of adapting GTA5 to Cityscapes. The first and second groups adopt VGG-16 and ResNet-101 networks, respectively.
	}
	\vspace{1mm}
	\label{table:gta5_all}
	\footnotesize
	\centering
	\renewcommand{\arraystretch}{1.2}
	\setlength{\tabcolsep}{2.4pt}
	\begin{tabular}{lcccccccccccccccccccc}
		\toprule
		
		& \multicolumn{20}{c}{GTA5 $\rightarrow$ Cityscapes} \\
		\midrule
		
		Method & \rotatebox{90}{road} & \rotatebox{90}{sidewalk} & \rotatebox{90}{building} & \rotatebox{90}{wall} & \rotatebox{90}{fence} & \rotatebox{90}{pole} & \rotatebox{90}{light} & \rotatebox{90}{sign} & \rotatebox{90}{veg} & \rotatebox{90}{terrain} & \rotatebox{90}{sky} & \rotatebox{90}{person} & \rotatebox{90}{rider} & \rotatebox{90}{car} & \rotatebox{90}{truck} & \rotatebox{90}{bus} & \rotatebox{90}{train} & \rotatebox{90}{mbike} & \rotatebox{90}{bike} & mIoU\\
		
		\midrule
		
		FCNs in the Wild~\cite{Hoffman_CoRR_2016} & 70.4 & 32.4 & 62.1 & 14.9 & 5.4 & 10.9 & 14.2 & 2.7 & 79.2 & 21.3 & 64.6 & 44.1 & 4.2 & 70.4 & 8.0 & 7.3 & 0.0 & 3.5 & 0.0 & 27.1 \\
		
		CDA~\cite{Zhang_ICCV_2017} & 74.9 & 22.0 & 71.7 & 6.0 & 11.9 & 8.4 & 16.3 & 11.1 & 75.7 & 13.3 & 66.5 & 38.0 & 9.3 & 55.2 & 18.8 & 18.9 & 0.0 & 16.8 & 14.6 & 28.9 \\
		
		ST~\cite{Zou_ECCV_2018} & 83.8 & 17.4 & 72.1 & 14.3 & 2.9 & 16.5 & 16.0 & 6.8 & \textbf{81.4} & 24.2 & 47.2 & 40.7 & 7.6 & 71.7 & 10.2 & 7.6 & 0.5 & 11.1 & 0.9 & 28.1 \\
		
		CBST~\cite{Zou_ECCV_2018} & 66.7 & 26.8 & 73.7 & 14.8 & 9.5 & \textbf{28.3} & 25.9 & 10.1 & 75.5 & 15.7 & 51.6 & 47.2 & 6.2 & 71.9 & 3.7  & 2.2 & \textbf{5.4} & \textbf{18.9} & \textbf{32.4} & 30.9 \\
		
		CyCADA~\cite{Hoffman_ICML_2018} & 83.5 & \textbf{38.3} & 76.4 & 20.6 & 16.5 & 22.2 & \textbf{26.2} & \textbf{21.9} & 80.4 & 28.7 & 65.7 & 49.4 & 4.2 & 74.6 & 16.0 & 26.6 & 2.0 & 8.0 & 0.0 & 34.8 \\
		
		
		Output Space~\cite{Tsai_CVPR_2018} & \textbf{87.3} & 29.8 & 78.6 & 21.1 & \textbf{18.2} & 22.5 & 21.5 & 11.0 & 79.7 & 29.6 & \textbf{71.3} & 46.8 & 6.5 & 80.1 & \textbf{23.0} & 26.9 & 0.0 & 10.6 & 0.3 & 35.0 \\
		
		Ours (VGG-16) & \textbf{87.3} & 35.7 & \textbf{79.5} & \textbf{32.0} & 14.5 & 21.5 & 24.8 & 13.7 & 80.4 & \textbf{32.0} & 70.5 & \textbf{50.5} & \textbf{16.9} & \textbf{81.0} & 20.8 & \textbf{28.1} & 4.1 & 15.5 & 4.1 & \textbf{37.5} \\
		
		\midrule
		Without Adaptation & 75.8 & 16.8 & 77.2 & 12.5 & 21.0 & 25.5 & 30.1 & 20.1 & 81.3 & 24.6 & 70.3 & 53.8 & 26.4 & 49.9 & 17.2 & 25.9 & 6.5 & 25.3 & \textbf{36.0} & 36.6 \\
		
		Feature Space~\cite{Tsai_CVPR_2018} & 83.7 & 27.6 & 75.5 & 20.3 & 19.9 & 27.4 & 28.3 & 27.4 & 79.0 & 28.4 & 70.1 & 55.1 & 20.2 & 72.9 & 22.5 & 35.7 & \textbf{8.3} & 20.6 & 23.0 & 39.3 \\
		
		Road~\cite{Chen_CVPR_2018} & 76.3 & 36.1 & 69.6 & 28.6 & 22.4 & \textbf{28.6} & 29.3 & 14.8 & 82.3 & \textbf{35.3} & 72.9 & 54.4 & 17.8 & 78.9 & 27.7 & 30.3 & 4.0 & 24.9 & 12.6 & 39.4 \\
		
		
		Output Space~\cite{Tsai_CVPR_2018} & 86.5 & 25.9 & 79.8 & 22.1 & 20.0 & 23.6 & 33.1 & 21.8 & 81.8 & 25.9 & 75.9 & 57.3 & 26.2 & 76.3 & 29.8 & 32.1 & 7.2 & \textbf{29.5} & 32.5 & 41.4 \\

		
		Ours (ResNet-101) & \textbf{92.3} & \textbf{51.9} & \textbf{82.1} & \textbf{29.2} & \textbf{25.1} & 24.5 & \textbf{33.8} & \textbf{33.0} & \textbf{82.4} & 32.8 & \textbf{82.2} & \textbf{58.6} & \textbf{27.2} & \textbf{84.3} & \textbf{33.4} & \textbf{46.3} & 2.2 & \textbf{29.5} & 32.3 & \textbf{46.5} \\
		\bottomrule
	\end{tabular}
	\vspace{3mm}
\end{table*}
\begin{table*} [t]
	\caption{
		Results of adapting SYNTHIA to Cityscapes. The first and second groups adopt VGG-16 and ResNet-101 networks, respectively. mIoU and mIoU$^\ast$ are averaged over 16 and 13 categories, respectively.
	}
	\vspace{1mm}
	\label{table:synthia}
	\footnotesize
	\centering
	\renewcommand{\arraystretch}{1.2}
	\setlength{\tabcolsep}{3pt}
	\begin{tabular}{lcccccccccccccccccc}
		\toprule
		
		& \multicolumn{18}{c}{SYNTHIA $\rightarrow$ Cityscapes} \\
		\midrule
		
		Method & \rotatebox{90}{road} & \rotatebox{90}{sidewalk} & \rotatebox{90}{building} & \rotatebox{90}{wall} & \rotatebox{90}{fence} & \rotatebox{90}{pole} & \rotatebox{90}{light} & \rotatebox{90}{sign} & \rotatebox{90}{veg} & \rotatebox{90}{sky} & \rotatebox{90}{person} & \rotatebox{90}{rider} & \rotatebox{90}{car} & \rotatebox{90}{bus} & \rotatebox{90}{mbike} & \rotatebox{90}{bike} & mIoU & mIoU$^\ast$ \\
		
		\midrule
		
		FCNs in the Wild~\cite{Hoffman_CoRR_2016} & 11.5 & 19.6 & 30.8 & \textbf{4.4} & 0.0 & 20.3 & 0.1 & \textbf{11.7} & 42.3 & 68.7 & 51.2 & 3.8 & 54.0 & 3.2 & 0.2 & 0.6 & 20.2 & 22.1 \\
		
		CDA~\cite{Zhang_ICCV_2017} & 65.2 & 26.1 & 74.9 & 0.1 & \textbf{0.5} & 10.7 & \textbf{3.7} & 3.0 & 76.1 & 70.6 & 47.1 & 8.2 & 43.2 & \textbf{20.7} & 0.7 & 13.1 & 29.0 & 34.8 \\
		
		Cross-City~\cite{Chen_ICCV_2017} & 62.7 & 25.6 & \textbf{78.3} & - & - & - & 1.2 & 5.4 & \textbf{81.3} & \textbf{81.0} & 37.4 & 6.4 & 63.5 & 16.1 & 1.2 & 4.6 & - & 35.7 \\
		
		ST~\cite{Zou_ECCV_2018} & 0.2 & 14.5 & 53.8 & 1.6 & 0.0 & 18.9 & 0.9 & 7.8 & 72.2 & 80.3 & \textbf{48.1} & 6.3 & 67.7 & 4.7 & 0.2 & 4.5 & 23.9 & 27.8 \\
		
		
		
		Output Space~\cite{Tsai_CVPR_2018} & \textbf{78.9} & 29.2 & 75.5 & - & - & - & 0.1 & 4.8 & 72.6 & 76.7 & 43.4 & 8.8 & 71.1 & 16.0 & 3.6 & 8.4 & - & 37.6 \\
		
		Ours (VGG-16) & 72.6 & \textbf{29.5} & 77.2 & 3.5 & 0.4 & \textbf{21.0} & 1.4 & 7.9 & 73.3 & 79.0 & 45.7 & \textbf{14.5} & \textbf{69.4} & 19.6 & \textbf{7.4} & \textbf{16.5} & \textbf{33.7} & \textbf{39.6} \\
		
		\midrule
		Without Adaptation & 55.6 & 23.8 & 74.6 & 9.2 & 0.2 & 24.4 & 6.1 & \textbf{12.1} & 74.8 & 79.0 & \textbf{55.3} & 19.1 & 39.6 & 23.3 & 13.7 & 25.0 & 33.5 & 38.6 \\
		
		Feature Space~\cite{Tsai_CVPR_2018} & 62.4 & 21.9 & 76.3 & \textbf{11.5} & 0.1 & 24.9 & \textbf{11.7} & 11.4 & 75.3 & 80.9 & 53.7 & 18.5 & 59.7 & 13.7 & 20.6 & 24.0 & 35.4 & 40.8 \\
		
		Output Space~\cite{Tsai_CVPR_2018} & 79.2 & 37.2 & \textbf{78.8} & 10.5 & 0.3 & 25.1 & 9.9 & 10.5 & \textbf{78.2} & 80.5 & 53.5 & 19.6 & 67.0 & 29.5 & \textbf{21.6} & 31.3 & 39.5 & 45.9 \\
		
		Ours (ResNet-101) & \textbf{82.4} & \textbf{38.0} & 78.6 & 8.7 & \textbf{0.6} & \textbf{26.0} & 3.9 & 11.1 & 75.5 & \textbf{84.6} & 53.5 & \textbf{21.6} & \textbf{71.4} & \textbf{32.6} & 19.3 & \textbf{31.7} & \textbf{40.0} & \textbf{46.5} \\
		
		\bottomrule
	\end{tabular}
\end{table*}
\subsection{Comparisons with State-of-the-art Methods}
%
We have validated that the proposed patch-level alignment is complementary to existing domain adaptation methods on semantic segmentation.
%
%
%
In the following, we compare our final model (Fu + Patch-Alignment) with state-of-the-art algorithms under various scenarios, including synthetic-to-real and cross-city cases.
%
%
\vspace{-3mm}
\paragraph{Synthetic-to-real Case.}
We first present experimental results for adapting GTA5 to Cityscapes in Table~\ref{table:gta5_all}.
We utilize two different architectures, i.e., VGG-16 and ResNet-101, and compare with state-of-the-art approaches via feature adaptation~\cite{Hoffman_CoRR_2016, Zhang_ICCV_2017}, pixel-level adaptation~\cite{Hoffman_ICML_2018}, pseudo label re-training~\cite{Zou_ECCV_2018} and output space alignment~\cite{Chen_CVPR_2018, Tsai_CVPR_2018}.
%
%
%
%
We show that the proposed framework improves over existing methods by 2.5\% and 5.1\% in mean IoU for two architectures, respectively.
%
%
In Table~\ref{table:synthia}, we present results for adapting SYNTHIA to Cityscapes and similar improvements are observed compared to state-of-the-arts.
%
%
%
%
In addition, we shows visual comparisons in Figure~\ref{fig: visual} and more results are presented in the supplementary material.
\vspace{-3mm}
\paragraph{Cross-city Case.}
Adapting between real images across different cities and conditions is an important scenario for practical applications.
We choose a challenging case where the weather condition is different (i.e., sunny v.s. rainy) in two cities by adapting Cityscapes to Oxford RobotCar.
The proposed framework achieves a mean IoU of 72.0$\%$ averaged on 9 categories, significantly improving the model without adaptation by 10.1$\%$.
To compare with the output space adaptation method~\cite{Tsai_CVPR_2018}, we run the authors' released code and obtain a mean IoU of 69.5$\%$, which is 2.5$\%$ lower than the proposed method.
Further results and comparisons are provided in the supplementary material.
%

%% file: conclusion.tex
\section{Conclusions}
%
In this paper, we present a domain adaptation method for structured output via patch-level alignment.
We propose to learn discriminative representations of patches by constructing a clustered space of the source patches and adopt an adversarial learning scheme to push the target patch distributions closer to the source ones.
With patch-level alignment, our method is complementary to various domain adaptation approaches and provides additional improvement.
We conduct extensive ablation studies and experiments to validate the effectiveness of the proposed method under numerous challenges on semantic segmentation, including synthetic-to-real and cross-city scenarios, and show that our approach performs favorably against existing algorithms.

%% file: supp.tex
\begin{table*} [!ht]
	\caption{
		Image and patch sizes for training and testing.
	}
	\vspace{1mm}
	\label{table:size}
	\small
	\centering
	\renewcommand{\arraystretch}{1.2}
	\setlength{\tabcolsep}{5.5pt}
	\begin{tabular}{lcccc}
		\toprule
		
		Dataset & Cityscapes & GTA5 & SYNTHIA & Oxford RobotCar \\
		\midrule
		Patch size for training & $32 \times 64$ & $36 \times 64$ & $38 \times 64$ & - \\
		
		Image size for training& $512 \times 1024$ & $720 \times 1280$ & $760 \times 1280$ & $960 \times 1280$ \\
		
		Image size for testing & $512 \times 1024$ & - & - & $960 \times 1280$ \\
		\bottomrule
	\end{tabular}
\end{table*}
\section{Training Details}
To train the model in an end-to-end manner, we randomly sample one image from each of the source and target domain (i.e., batch size as 1) in a training iteration.
Then we follow the optimization strategy as described in Section 3.3 of the main paper.
Table~\ref{table:size} shows the image and patch sizes during training and testing. Note that, the aspect ratio of the image is always maintained (i.e., no cropping) and then the image is down-sampled to the size as in the table.
\begin{table*} [!t]
	\caption{
		Results of adapting Cityscapes to Oxford RobotCar.
	}
	\vspace{1mm}
	\label{table:oxford}
	\centering
	\renewcommand{\arraystretch}{1.2}
	\setlength{\tabcolsep}{5pt}
	\begin{tabular}{lccccccccccc}
		\toprule
		
		& \multicolumn{10}{c}{Cityscapes $\rightarrow$ Oxford RobotCar} \\
		\midrule
		
		Method & \rotatebox{90}{road} & \rotatebox{90}{sidewalk} & \rotatebox{90}{building} & \rotatebox{90}{light} & \rotatebox{90}{sign} & \rotatebox{90}{sky} & \rotatebox{90}{person} & \rotatebox{90}{automobile} & \rotatebox{90}{two-wheel} & mIoU\\
		
		\midrule
		
		Without Adaptation & 79.2 & 49.3 & 73.1 & 55.6 & 37.3 & 36.1 & 54.0 & 81.3 & 49.7 & 61.9 \\
		
		Output Space~\cite{Tsai_CVPR_2018}  & 95.1 & 64.0 & 75.7 & 61.3 & 35.5 & 63.9 & 58.1 & 84.6 & 57.0 & 69.5 \\
		
		Ours & 94.4 & 63.5 & 82.0 & 61.3 & 36.0 & 76.4 & 61.0 & 86.5 & 58.6 & 72.0 \\
		
		\bottomrule
	\end{tabular}
\end{table*}
\section{Relation to Entropy Minimization}
Entropy minimization~\cite{Grandvalet_NIPS_2014} can be used as a loss in our model to push the target feature representation $F_t$ to one of the source clusters.
To add this regularization, we replace the adversarial loss on the patch level with an entropy loss as in~\cite{long2016unsupervised}, where the entropy loss $\mathcal{L}_{en} = \sum_{u,v} \sum_{k} H(\sigma(F_t/\tau))^{(u,v,k)}$, $H$ is the information entropy function, $\sigma$ is the softmax function, and $\tau$ is the temperature of the softmax.
The model with adding this entropy regularization achieves the IoU as 41.9\%, that is lower than the proposed patch-level adversarial alignment as 43.2\%.
The reason is that, different from the entropy minimization approach that does not use the source distribution as the guidance, our model learns discriminative representations for the target patches by pushing them closer to the source distribution in the clustered space guided by the annotated labels.
\section{More Ablation Study on Clustered Space}
%

To validate the effectiveness of the $\mathbf{H}$ module, we conduct an experiment on GTA5-to-Cityscapes that directly computes category histograms from the segmentation output and then perform alignment.
This implementation achieves an IoU $0.7\%$ lower than our method as $41.3 \%$. A possible reason is that we use the $\mathbf{H}$ module that involves learnable parameters to estimate K-means memberships, whereas directly computing category histograms would solely rely on updating the segmentation network $\mathbf{G}$, which causes more training difficulty.
\section{More Details on Pseudo Label Re-training}
We use the official implementation of \cite{Hung_BMVC_2018} provided by the authors. In this case, we consider our target samples as unlabeled data used in \cite{Hung_BMVC_2018} under the semi-supervised setting.
The same discriminator in the output space and the loss function are then adopted as in \cite{Hung_BMVC_2018}.
\section{Result of Cityscapes-to-Oxford}
\begin{figure*}[h!]
	\centering
	\begin{tabular}
		{@{\hspace{0mm}}c@{\hspace{1mm}} @{\hspace{0mm}}c@{\hspace{1mm}}
			@{\hspace{0mm}}c@{\hspace{1mm}} @{\hspace{0mm}}c@{\hspace{1mm}}
			@{\hspace{0mm}}c@{\hspace{0mm}}
		}
		\includegraphics[width=0.19\linewidth]{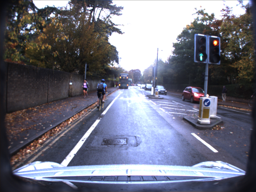} &
		\includegraphics[width=0.19\linewidth]{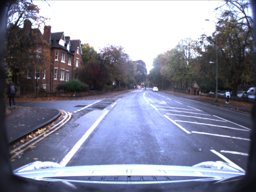} &
		\includegraphics[width=0.19\linewidth]{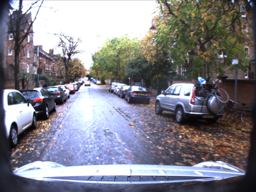} &
		\includegraphics[width=0.19\linewidth]{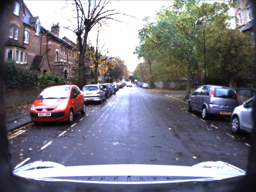} &
		\includegraphics[width=0.19\linewidth]{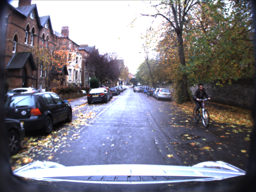} \\
		
		\includegraphics[width=0.19\linewidth]{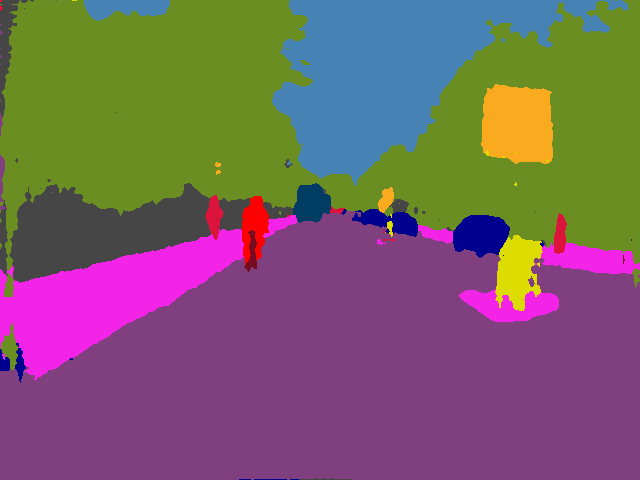} &
		\includegraphics[width=0.19\linewidth]{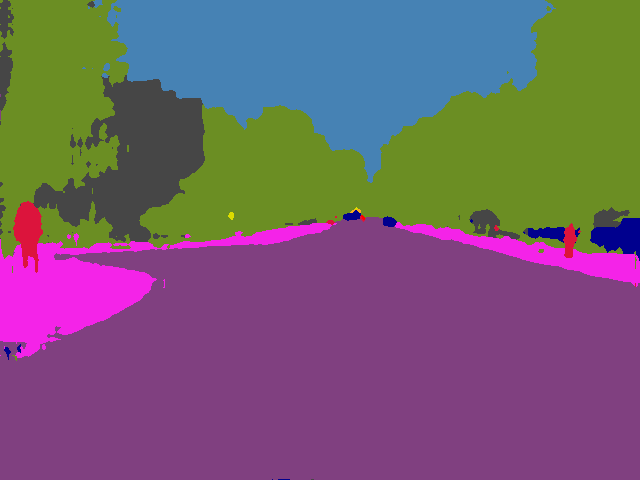} &
		\includegraphics[width=0.19\linewidth]{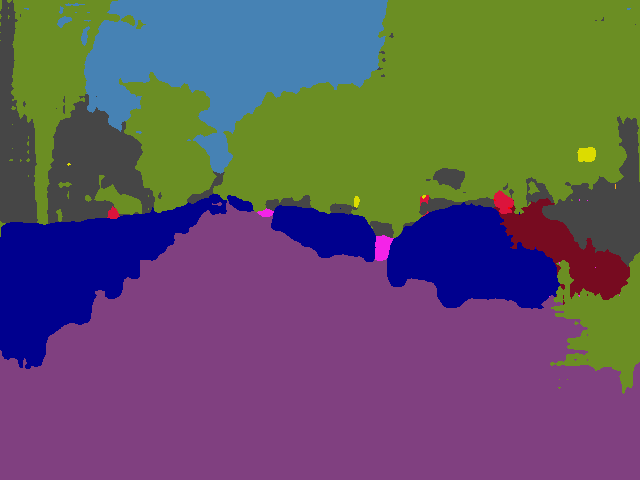} &
		\includegraphics[width=0.19\linewidth]{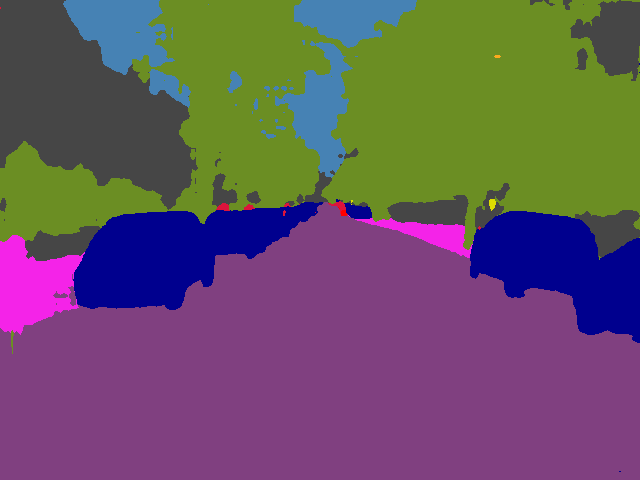} &
		\includegraphics[width=0.19\linewidth]{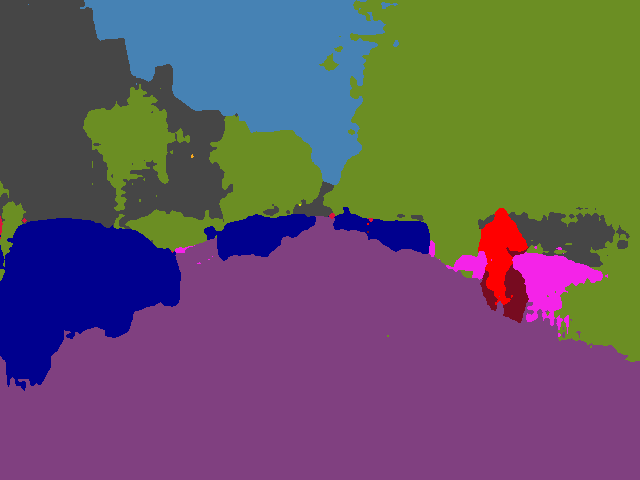} \\
		
		\\
		
		\includegraphics[width=0.19\linewidth]{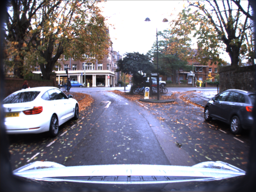} &
		\includegraphics[width=0.19\linewidth]{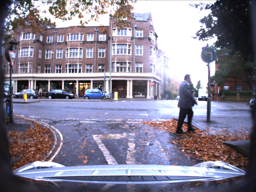} &
		\includegraphics[width=0.19\linewidth]{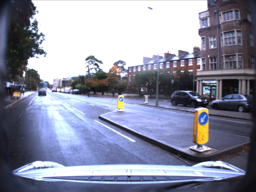} &
		\includegraphics[width=0.19\linewidth]{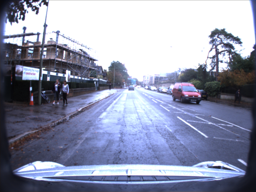} &
		\includegraphics[width=0.19\linewidth]{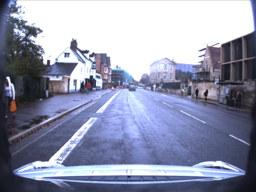} \\
		
		\includegraphics[width=0.19\linewidth]{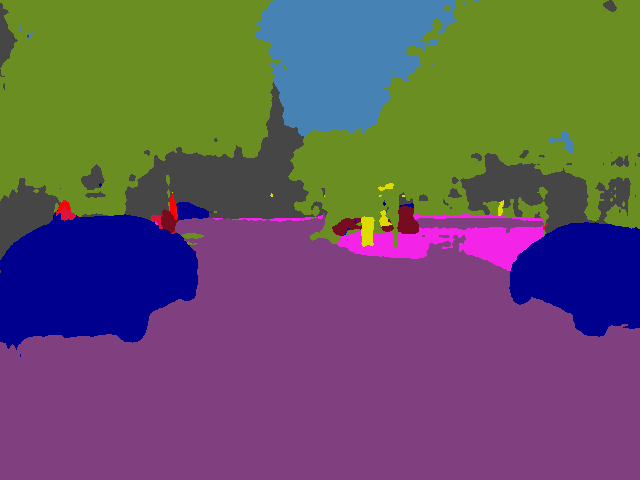} &
		\includegraphics[width=0.19\linewidth]{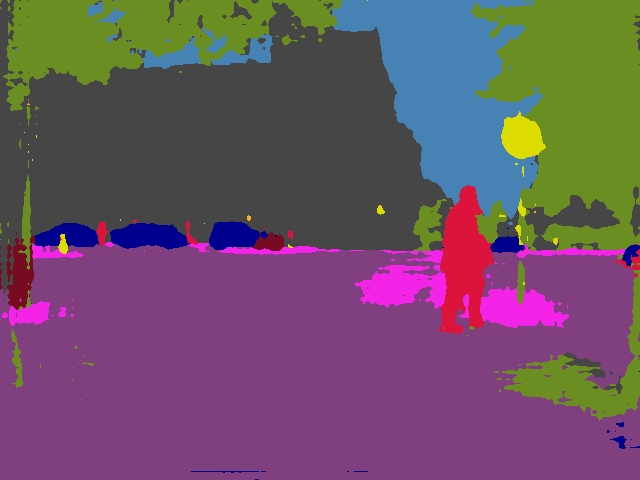} &
		\includegraphics[width=0.19\linewidth]{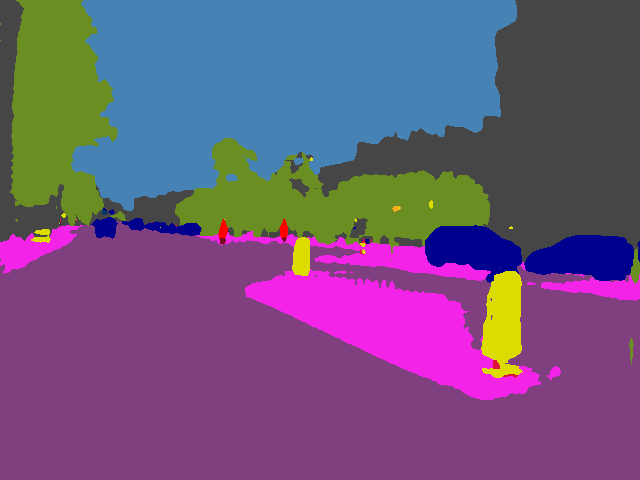} &
		\includegraphics[width=0.19\linewidth]{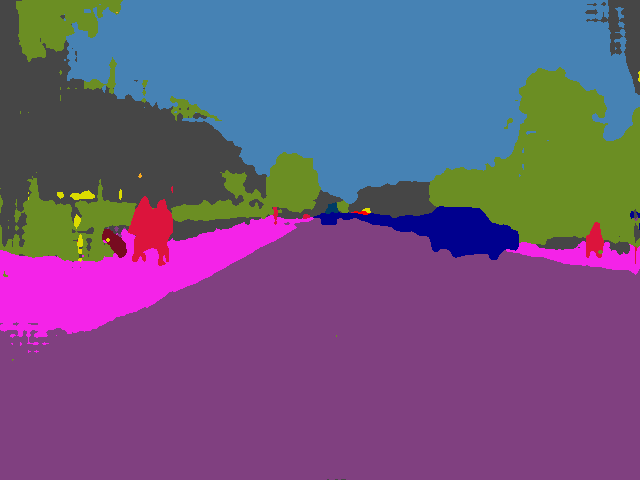} &
		\includegraphics[width=0.19\linewidth]{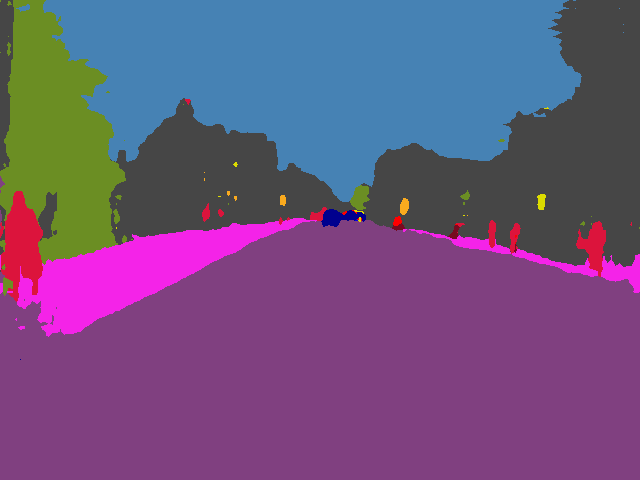} \\

	\end{tabular}
	\caption{Example results of adapted segmentation for the Cityscapes-to-OxfordRobotCar setting. We sequentially show images in a video and their adapted segmentations generated by our method.
	}
	\label{fig:oxford_1}
\end{figure*}

\begin{figure*}[h!]
	\centering
	\begin{tabular}
		{@{\hspace{0mm}}c@{\hspace{1mm}} @{\hspace{0mm}}c@{\hspace{1mm}}
			@{\hspace{0mm}}c@{\hspace{1mm}} @{\hspace{0mm}}c@{\hspace{1mm}}
			@{\hspace{0mm}}c@{\hspace{0mm}}
		}
		\includegraphics[width=0.19\linewidth]{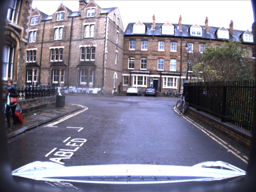} &
		\includegraphics[width=0.19\linewidth]{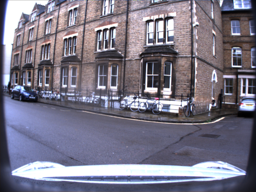} &
		\includegraphics[width=0.19\linewidth]{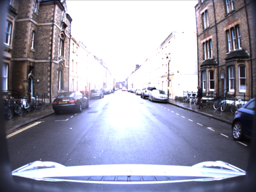} &
		\includegraphics[width=0.19\linewidth]{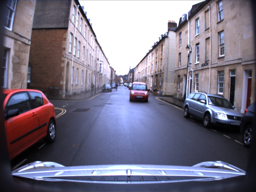} &
		\includegraphics[width=0.19\linewidth]{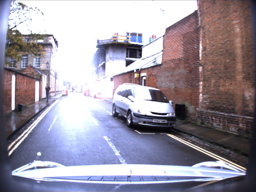} \\
		
		\includegraphics[width=0.19\linewidth]{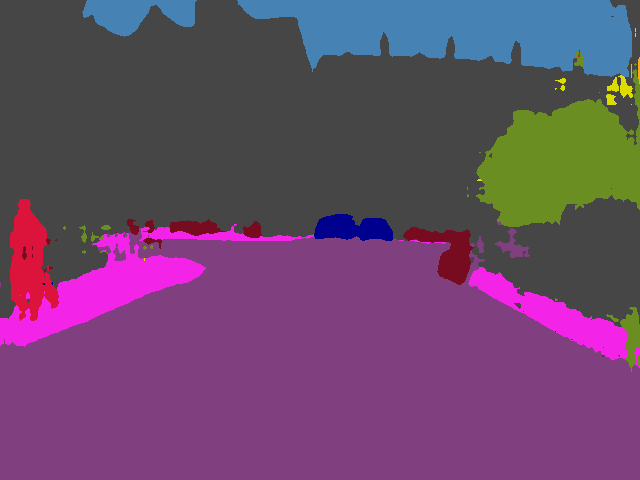} &
		\includegraphics[width=0.19\linewidth]{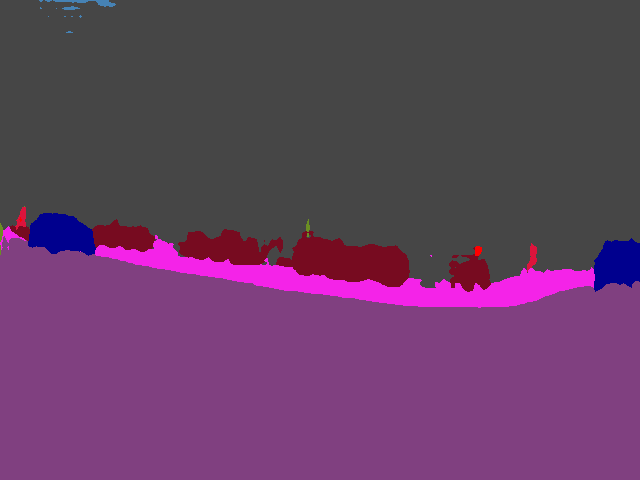} &
		\includegraphics[width=0.19\linewidth]{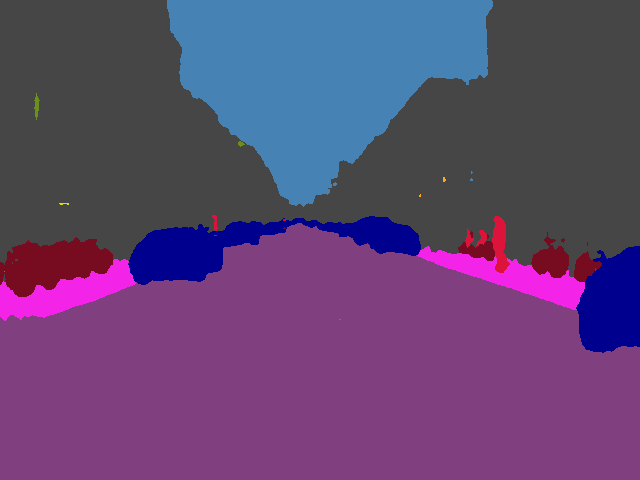} &
		\includegraphics[width=0.19\linewidth]{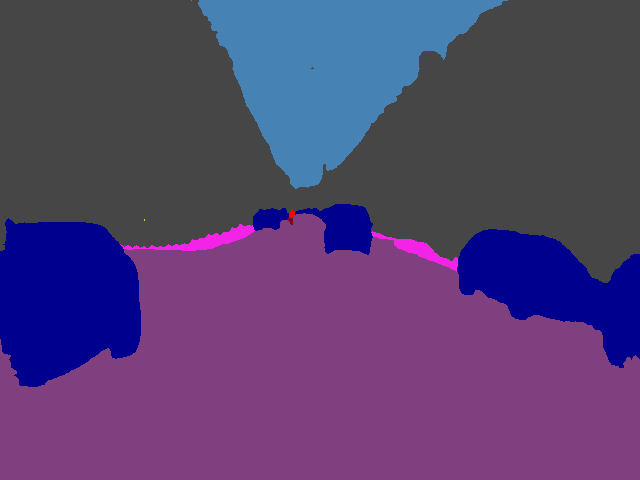} &
		\includegraphics[width=0.19\linewidth]{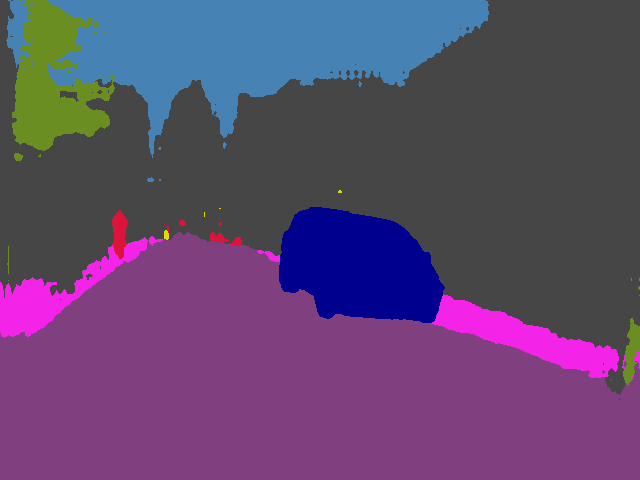} \\
		
		\\
		
		\includegraphics[width=0.19\linewidth]{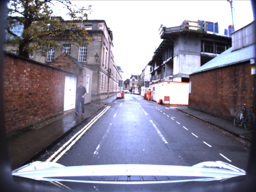} &
		\includegraphics[width=0.19\linewidth]{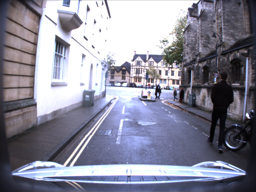} &
		\includegraphics[width=0.19\linewidth]{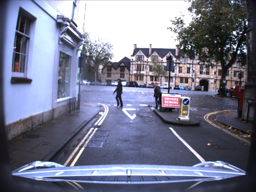} &
		\includegraphics[width=0.19\linewidth]{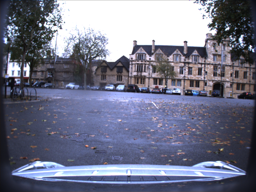} &
		\includegraphics[width=0.19\linewidth]{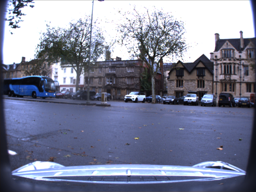} \\
		
		\includegraphics[width=0.19\linewidth]{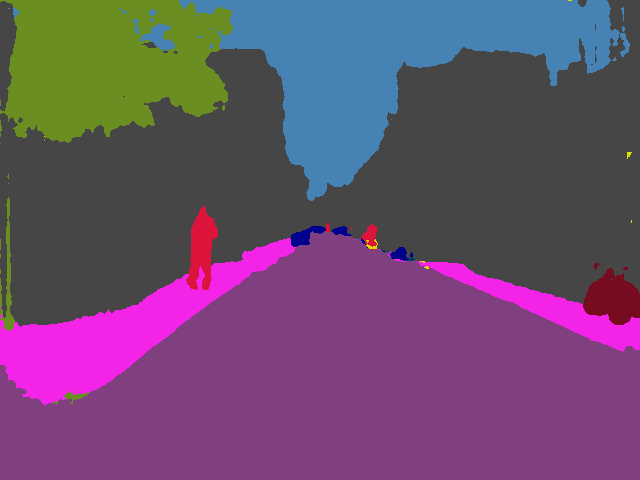} &
		\includegraphics[width=0.19\linewidth]{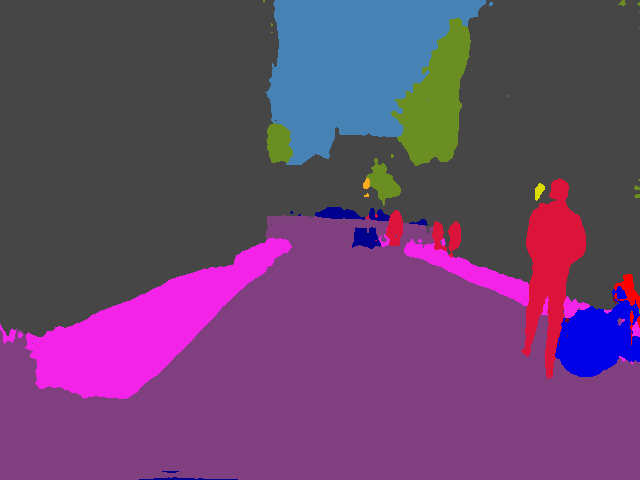} &
		\includegraphics[width=0.19\linewidth]{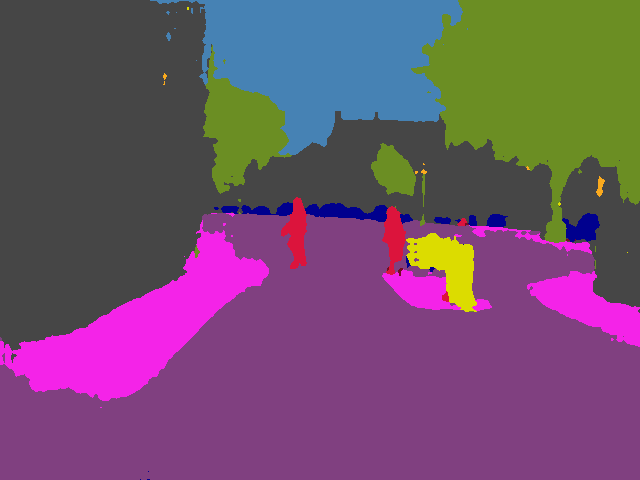} &
		\includegraphics[width=0.19\linewidth]{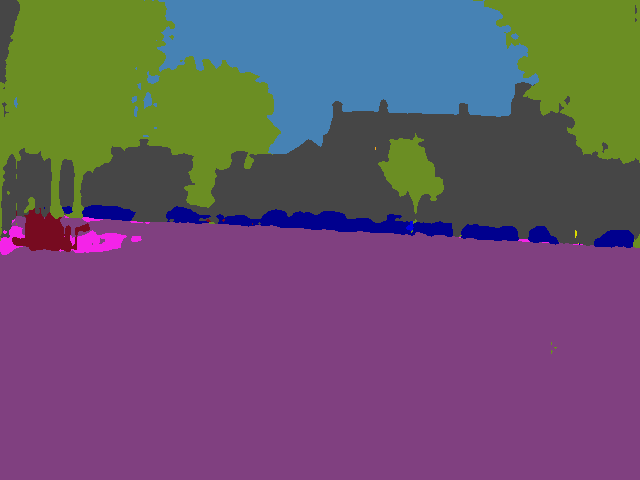} &
		\includegraphics[width=0.19\linewidth]{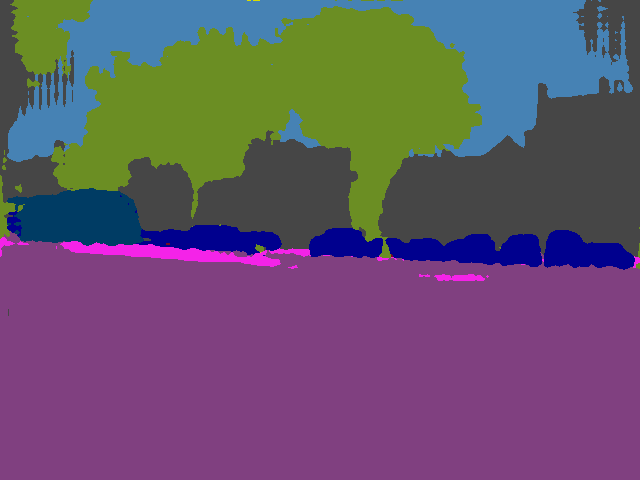} \\
		
	\end{tabular}
	\caption{Example results of adapted segmentation for the Cityscapes-to-OxfordRobotCar setting. We sequentially show images in a video and their adapted segmentations generated by our method.
	}
	\label{fig:oxford_2}

\end{figure*}

In Table~\ref{table:oxford}, we present the complete result for adapting Cityscapes (sunny condition) to Oxford RobotCar (rainy scene).
We compare the proposed method with the model without adaptation and the output space adaptation approach~\cite{Tsai_CVPR_2018}.
%
More qualitative results are provided in Figure~\ref{fig:oxford_1} and \ref{fig:oxford_2}.
\section{Qualitative Comparisons}
\begin{figure*}[t]
	\centering
	\includegraphics[width=1.0\linewidth]{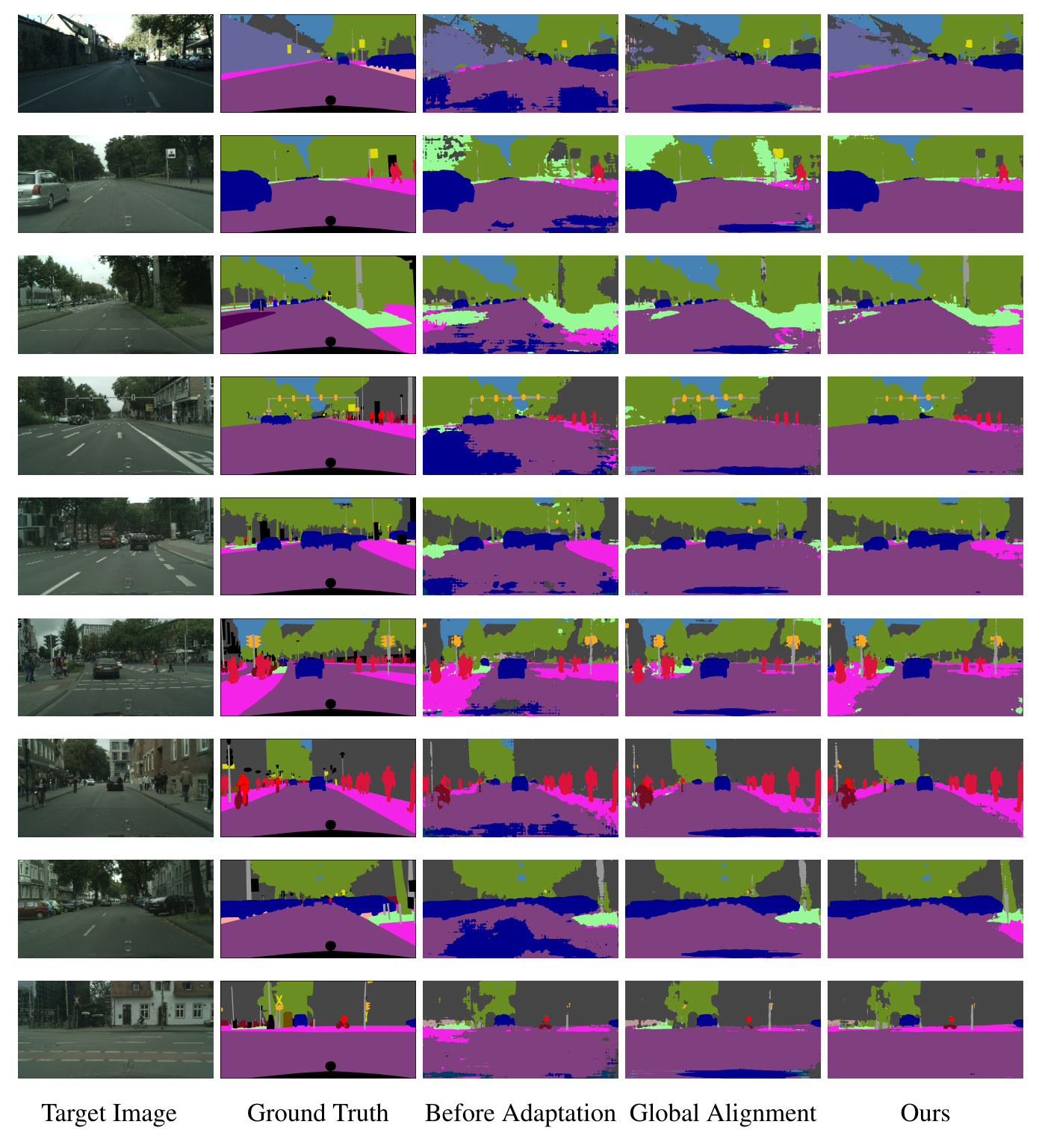}\\
	\caption{Example results of adapted segmentation for the GTA5-to-Cityscapes setting. For each target image, we show results before adaptation, output space adaptation~\cite{Tsai_CVPR_2018}, and the proposed method.
	}
	\label{fig:gta_1}
\end{figure*}

\begin{figure*}[t]
	\centering
	\includegraphics[width=1.0\linewidth]{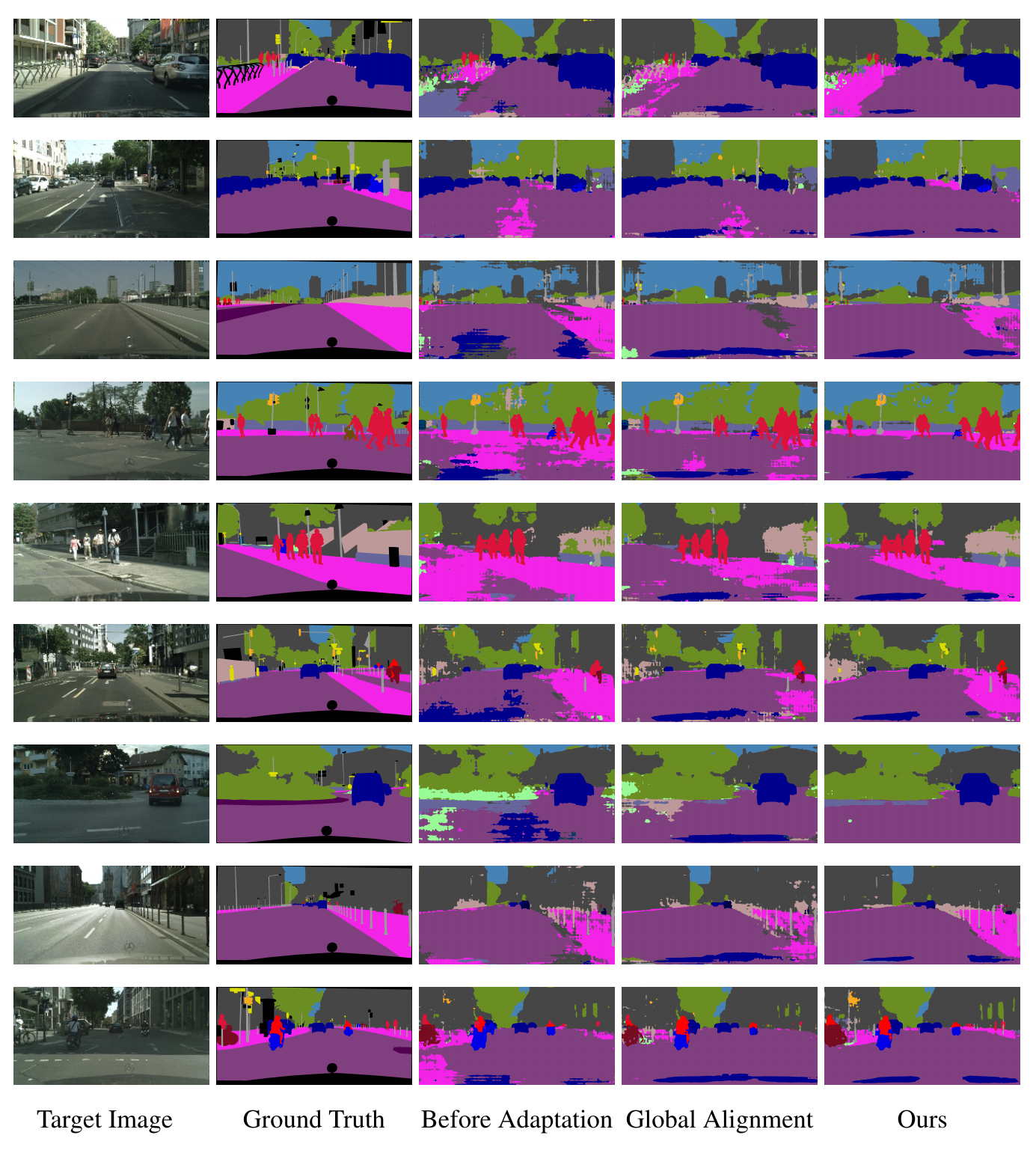}\\
	\caption{Example results of adapted segmentation for the GTA5-to-Cityscapes setting. For each target image, we show results before adaptation, output space adaptation~\cite{Tsai_CVPR_2018}, and the proposed method.
	}
	\label{fig:gta_2}
\end{figure*}

\begin{figure*}[t]
	\centering
	\includegraphics[width=1.0\linewidth]{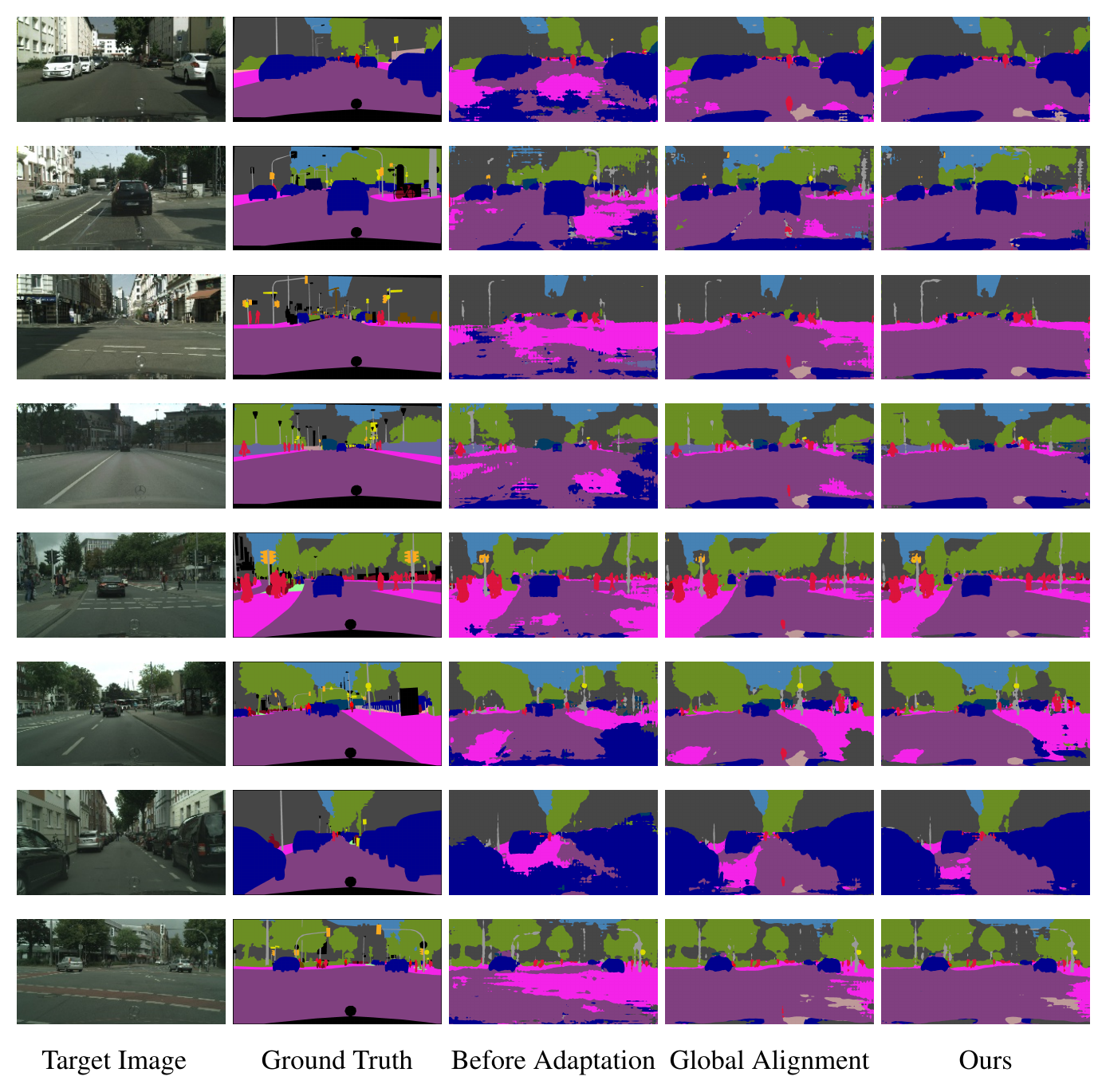}\\
	\caption{Example results of adapted segmentation for the SYNTHIA-to-Cityscapes setting. For each target image, we show results before adaptation, output space adaptation~\cite{Tsai_CVPR_2018}, and the proposed method.
	}
	\label{fig:synthia}
\end{figure*}

We provide more visual comparisons for GTA5-to-Cityscapes and SYNTHIA-to-Cityscapes scenarios from Figure~\ref{fig:gta_1} to Figure~\ref{fig:synthia}.
In each row, we present the results of the model without adaptation, output space adaptation~\cite{Tsai_CVPR_2018}, and the proposed method.
We show that our approach often yields better segmentation outputs with more details and produces less noisy regions.
